\pdfoutput=1

\documentclass[11pt]{article}

\usepackage[final]{acl}

\usepackage{times}
\usepackage{latexsym}
\usepackage{titling}

\usepackage[T1]{fontenc}

\usepackage[utf8]{inputenc}

\usepackage{ragged2e}
\usepackage{multirow}
\usepackage{booktabs}
\usepackage{amsmath}
\usepackage{inconsolata}

\usepackage{graphicx}
\usepackage{kotex}
\usepackage[verbose]{microtype}
\title{Benchmarking Foundation Models on Exceptional Cases: Dataset Creation and Validation}

\author{Suho Kang, Jungyang Park,  Joonseo Ha, SoMin Kim, JinHyeong Kim, Subeen Park, Kyungwoo Song\\
\\
Yonsei Universiy}
\begin{document}
\maketitle
\begin{abstract}
Foundation models (FMs) have achieved significant success across various tasks, leading to research on benchmarks for reasoning abilities. 
However, there is a lack of studies on FMs performance in exceptional scenarios, which we define as out-of-distribution (OOD) reasoning tasks. This paper is the first to address these cases, developing a novel dataset for evaluation of FMs across multiple modalities, including graphic novels, calligraphy, news articles, and lyrics. It includes tasks for instance classification, character recognition, token prediction, and text generation. The paper also proposes prompt engineering techniques like Chain-of-Thought (CoT) and CoT+Few-Shot to enhance performance. Validation of FMs using various methods revealed improvements. The code repository is accessible at: \href{https://github.com/MLAI-Yonsei/ExceptionalBenchmark}{https://github.com/MLAI-Yonsei/ExceptionalBenchmark}
\end{abstract}

\section{Introduction}
Recent studies \citep{sap2019atomic, speer2017conceptnet, talmor2018commonsenseqa} have focused on assessing the commonsense reasoning capabilities 
of foundation models (FMs) \citep{achiam2023gpt, team2023gemini}. As a result, current FMs have achieved remarkable progress, demonstrating high performance across various tasks \citep{cherian2023deep, wang2018glue, wang2019superglue}. However, there are situations where FMs struggle to determine reasoning. Despite the development of various datasets \citep{yue2023mmmu, zellers2019hellaswag, lin2024wildbench}, There is a need for more diverse datasets that encompass less common scenarios. We characterize these scenarios as \textbf{exceptional cases}, referring to situations that contravene commonsense knowledge \citep{sap2019atomic, speer2017conceptnet}. Consequently, We define an exceptional case in a reasoning task as one that is out-of-distribution (OOD). In essence, the joint probability distribution of Exceptional Cases (\(P_{te}(x, y)\)) differs from the joint probability distribution (\(P_{tr}(x, y)\)) of the training dataset that Foundation Models (FMs) have learned. This can be expressed as follows:
\vspace{-1.0em}
\[P_{tr}(x, y) \neq P_{te}(x, y)\]
This discrepancy arises from differences in one or more of the following distributions: \(P_{tr}(x) \neq P_{te}(x)\), \(P_{tr}(y) \neq P_{te}(y)\), \(P_{tr}(y|x) \neq P_{te}(y|x)\) \citep{yang2024generalized}. 
The datasets can be classified accordingly.
\begin{center}
\(P_{tr}(x) \neq P_{te}(x)\) (Graphic Novels, Calligraphy)\end{center}
The graphic novel contains strong cartoonish storylines that FMs have rarely encountered before. Additionally, calligraphy characters are artistically rendered, deviating from the standard forms that FMs seldom encounter in their training datasets.
\begin{center}
\(P_{tr}(y) \neq P_{te}(y)\) (Lyrics)\end{center}
The task involving lyrics evaluates whether FMs can accurately complete masked segments. Those that BERT \citep{devlin2018bert} failed to predict are designated as exceptional cases, representing scenarios FMs rarely encounter.

\begin{center}
\(P_{tr}(y|x) \neq P_{te}(y|x)\) (Onion, Not The Onion)\end{center}
In the case of Onion's plausible fake news and Not the Onion's real news that seem fake, the classification accuracy is lower because they have a different nature from the News that FMs have learned.

\begin{figure}[ht]
    \centerline{\includegraphics[width=\columnwidth]{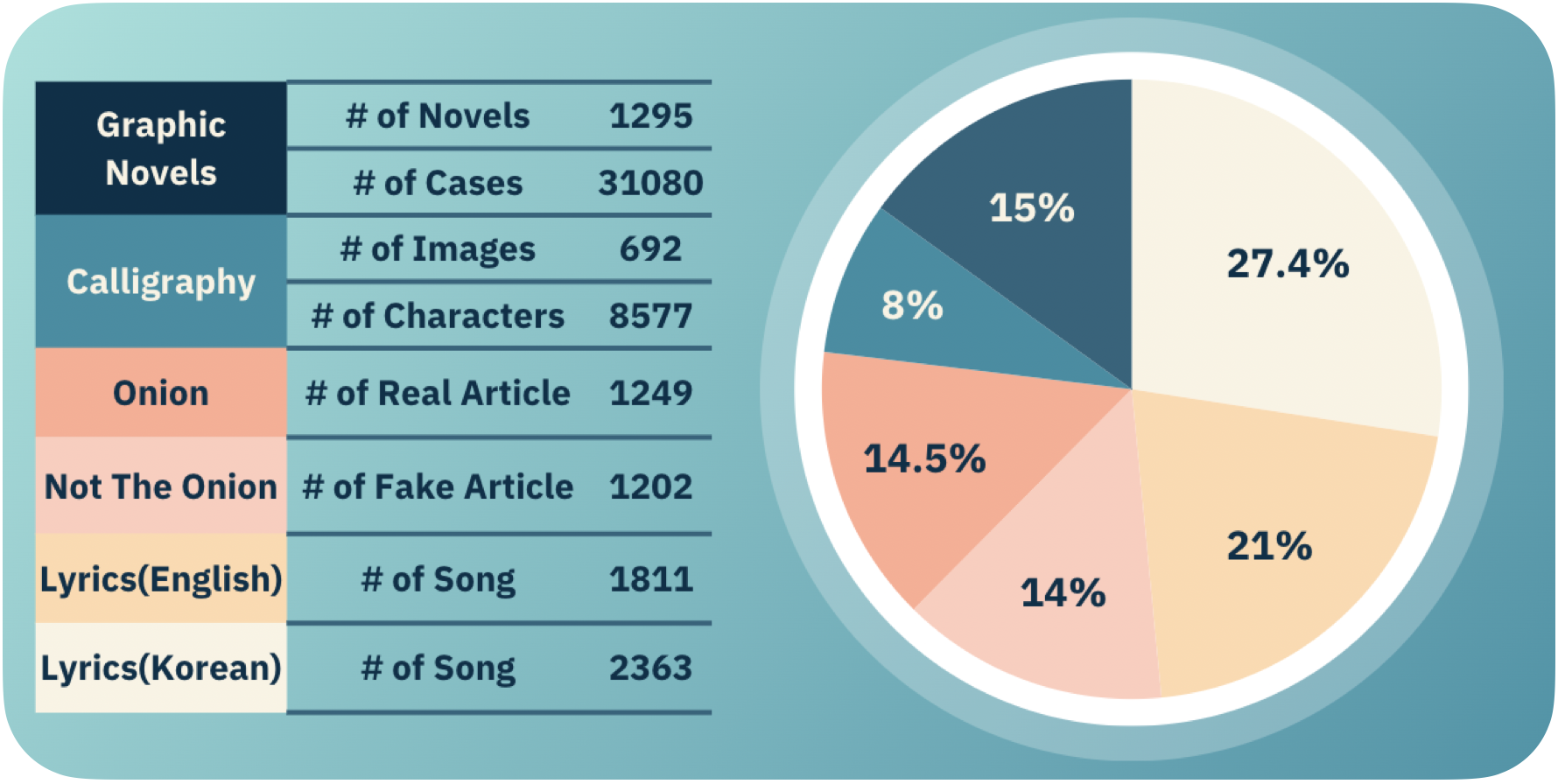}}
    \vspace{-5pt} 
    \caption{Distribution of Exceptional Cases Dataset and summary of four distinct datasets and their subsections.
    }
    \label{fig:piechart} 
\vspace{-1.0em}
\end{figure}
\section{Experiments and Results}
We designed experiments using four different datasets as shown in Figure \ref{fig:piechart}. That feature various characters with multi types of tasks such as instance recognition, text generation, token prediction, and character recognition. In the experiments for all four datasets, we conducted all experimental tasks using GPT-4o, Gemini-1.5-pro \citep{achiam2023gpt, team2023gemini}. Also, we employed three prompt styles—Zero-Shot \citep{kojima2022large}, Chain of Thought (CoT) \citep{wei2022chain}, and CoT+Few-Shot \citep{brown2020language}—to investigate how the accuracy of responses varies. The API temperature setting is regulated to 0 in GPT-4o, 0.01 in Gemini-1.5-pro to ensure consistent results.\\
\textbf{Experiments:} 
The Graphic Novels feature a random shuffle task where four input images are shuffled by code before being presented to the FMs as prompts. The FMs are then required to determine the correct order of the images. The Calligraphy features OCR tasks for transcribing Korean calligraphy. We initially planned to create an English Calligraphy dataset, but it is no longer considered an exceptional case since FMs have achieved high accuracy on it. For the WordArt \citep{shi2023exploring}, an English calligraphy dataset, GPT-4's accuracy is 60.20\%, but it increases to 77.61\% when evaluated on GPT-4o. The Onion, Not The Onion task comprises a binary classification task, where '0' corresponds to fake news and '1' to real news. The Lyrics dataset includes an infilling task, where the model predicts tokens for masked parts that BERT has identified as exceptional cases. Additionally, it features two more tasks: genre detection and song description generation.
\begin{table}[ht]\vspace{-0.5em}
\caption{Result(\%) of the random shuffle task}
\label{tab:graphic-result}\vspace{-0.5em}
\resizebox{\columnwidth}{!}{
\begin{tabular}{cccc}
\hline
\textbf{Acc.(\%)}       & \textbf{Zero-Shot} & \textbf{CoT}   & \textbf{CoT+Few-Shot} \\ \hline
\textbf{Claude-3.5-Sonnet} & 44.69     & 44.75 & 49.92        \\ \hline
\textbf{Gemini-1.5-Pro} & 51.41     & 52.45 & 52.51        \\ \hline
\textbf{GPT-4o}   & 63.80     & 63.88 & 64.63        \\ \hline
\end{tabular}
}\vspace{-1.0em}
\end{table}\newline
\textbf{Results:} In the task involving graphic novels, both baseline models demonstrated poor performance across all prompt styles, as shown in Table \ref{tab:graphic-result}. We analyzed the factors behind these poor results, suspecting that image style might hinder the FMs' reasoning. To test this, we prompted the models with a single image. The FMs gave detailed descriptions, identifying characters, actions, and hypothesizing thoughts. We then evaluated their accuracy with daily life narratives using everyday scenario images \citep{huang2016visual}, and the FMs responded well, showing adequate reasoning about storylines.

\begin{table}[ht]\vspace{-0.5em}
\caption{The results(\%) of the Korean Calligraphy OCR task indicate that the overall OCR capabilities of FMs are limited. GPT-4o exhibited superior performance, largely due to its enhanced ability to accurately detect spacing (’ ’) compared to other models.
}
\label{tab:calligraphy-result}\vspace{-0.5em}
\resizebox{\columnwidth}{!}{
\begin{tabular}{ccccc}
\hline
                             & \textbf{Model}          & \textbf{Zero-Shot} & \textbf{CoT}   & \textbf{CoT+Few-Shot} \\ \hline
\multirow{2}{*}{\textbf{Acc.(\%)(↑)}} & \textbf{Claude-3.5-Sonnet} & 14.42      & 26.40  & 32.20         \\ \cline{2-5}
                             & \textbf{Gemini-1.5-Pro} & 17.55      & 18.50  & 20.20         \\ \cline{2-5} 
                             & \textbf{GPT-4o}         & 53.43     & 61.54 & 61.86        \\ \hline
\multirow{2}{*}{\textbf{WER(\%)(↓)}}  & \textbf{Claude-3.5-Sonnet} & 81.39    & 74.00 & 65.62        \\ \cline{2-5}
                             & \textbf{Gemini-1.5-Pro} & 90.52    & 89.55 & 88.45        \\ \cline{2-5}
                             & \textbf{GPT-4o}         & 64.41     & 45.81 & 45.39        \\ \hline
\multirow{2}{*}{\textbf{CER(\%)(↓)}}  & \textbf{Claude-3.5-Sonnet} & 77.65     & 77.21 & 76.15        \\ \cline{2-5} 
                             & \textbf{Gemini-1.5-Pro} & 74.04     & 71.85 & 69.55        \\ \cline{2-5} 
                             & \textbf{GPT-4o}         & 32.64     & 24.73 & 22.55        \\ \hline
\end{tabular}
}\vspace{-1.0em}
\end{table}
In the Calligraphy OCR task, although the results were poor, FMs showed a tendency to use context for reasoning, suggesting that they were leveraging the relationships between words and characters. This tendency increased progressively across Zero-Shot, CoT, and CoT+Few-Shot approaches. However, despite this promising approach, it was ultimately unsuccessful, as performance declined when evaluated at the character-level, word-level, and overall meaning as shown in Table \ref{tab:calligraphy-result}.
\begin{table}[ht]\vspace{-0.5em}
\caption{Accuracy tended to decline with shorter articles. To investigate this trend in more detail, we divided the dataset into five sections based on article length, with Q1 representing the shortest and Q5 the longest articles. In Not The Onion, FMs often misclassified real short articles as fake due to the common association between shorter length and fake news. This tendency led to a noticeable drop in performance. Despite the task being a binary classification, Claude-3.5-Sonnet achieved only 69.30\% accuracy.}
\label{tab:onion-result}\vspace{-0.5em}
\resizebox{\columnwidth}{!}{
\begin{tabular}{cccccccc}
\hline
\multicolumn{2}{c}{\textbf{Length of Article}} & \textbf{Model}  & \textbf{Q1}    & \textbf{Q2}    & \textbf{Q3}     & \textbf{Q4}     & \textbf{Q5}     \\ \hline
\multirow{2}{*}{\textbf{Not The Onion}} & \multirow{2}{*}{\textbf{Acc.(\%)}} & \textbf{Claude-3.5-Sonnet} & 69.30 & 80.59 & 73.63  & 79.60  & 89.05  \\ \cline{3-8} 
                  &                   & \textbf{Gemini-1.5-Pro} & 67.35 & 80.20 & 89.58  & 92.18  & 91.66  \\ \cline{3-8}   
                  &                   & \textbf{GPT-4o} & 84.23 & 90.42 & 91.25  & 96.25  & 96.68  \\ \hline
\multirow{2}{*}{\textbf{Onion}}         & \multirow{2}{*}{\textbf{Acc.(\%)}} & \textbf{Claude-3.5-Sonnet} & 78.22 & 83.87 & 98.79 & 99.59 & 99.59 \\ \cline{3-8} 
                  &                   & \textbf{Gemini-1.5-Pro} & 92.98 & 99.12 & 100.00 & 100.00 & 100.00 \\ \cline{3-8} 
                  &                   & \textbf{GPT-4o} & 91.20 & 96.80 & 100.00 & 100.00 & 100.00 \\ \hline
\end{tabular}
}\vspace{-0.5em}
\end{table}
\newline Overall, the Onion, Not The Onion task showed strong results, but we observed that accuracy tended to decrease with shorter article lengths as shown Table \ref{tab:onion-result}. This suggests that FMs struggled to fully grasp the content of the news, likely due to its nuanced nature, often misclassifying shorter real articles as fake due to the common link between short length and fake news. 

%
\begin{table}[ht]\vspace{-0.5em}
\caption{The poor results for the lyrics infilling task indicate that FMs struggle with predicting tokens involving irregular and complex sentence structures and words.}
\label{tab:lyrics-result}\vspace{-0.5em}
\resizebox{\columnwidth}{!}{
\begin{tabular}{cccccc}
\hline
\multicolumn{2}{c}{\textbf{Infilling Result}}                        & \textbf{Baseline Model} & \textbf{Zero-shot} & \textbf{CoT}   & \textbf{CoT+Few-shot} \\ \hline
\multirow{2}{*}{\textbf{English}} & \multirow{2}{*}{\textbf{BERT Score(F1)}} & \textbf{Gemini-1.5-Pro} & 0.613     & 0.616 & 0.643        \\ \cline{3-6} 
 &  & \textbf{GPT-4o} & 0.611 & 0.632 & 0.653 \\ \hline
\multirow{2}{*}{\textbf{Korean}}  & \multirow{2}{*}{\textbf{BERT Score(F1)}} & \textbf{Gemini-1.5-Pro} & 0.032     & 0.155 & 0.324        \\ \cline{3-6} 
 &  & \textbf{GPT-4o} & 0.398 & 0.447 & 0.463 \\ \hline
\end{tabular}
}\vspace{-1.0em}
\end{table}
In the infilling task, the FMs exhibited poor performance. It is evident that the FMs struggle to predict the masked portions of lyrics classified as exceptional cases by BERT. Additionally, within the Korean dataset, we observe a significant performance degradation in Gemini-1.5-Pro compared to GPT-4o. In the English dataset, FMs refused to respond to songs released before the cut-off date, so the evaluation focused on music released after the cut-off date.
\section{Limitation}
This paper pioneers research into exceptional cases, which we define as out-of-distribution (OOD) scenarios in reasoning tasks. It aims to explore how FMs, recognized for their high performance across various domains, can address situations they typically struggle with, thereby advancing towards human-like reasoning. To this end, the study develops datasets encompassing diverse modalities, including image-only, text-only, and multimodal combinations. However, current research still lacks coverage of exceptional cases such as audio data \citep{yang2024air}. Future studies should establish benchmarks for exceptional cases in these and other unaddressed domains, defining appropriate tasks for their evaluation. We have only addressed English and Korean languages, leaving third-country languages unexplored and providing opportunities for further expansion. We utilized FMs ensure precise grammar and word usage.

\section*{Acknowledgements}
This work was supported by the National Research Foundation of Korea(NRF) grant funded by the Korea government(MSIT)(RS-2024-00457216, 2022R1A4A3033874).

\bibliography{2page/main}

\newpage
\appendix


\twocolumn[{%
    \centering
    \Large \textbf{Supplementary Material for Benchmarking Foundation Models on Exceptional Cases: Dataset Creation and Validation}\vspace{2.0em} \par
}]



\section{Graphic Novels}
\subsection{Task Details}
 \label{appendix: A.1}
\begin{figure}[ht]
    \centerline{\includegraphics[width=\columnwidth]{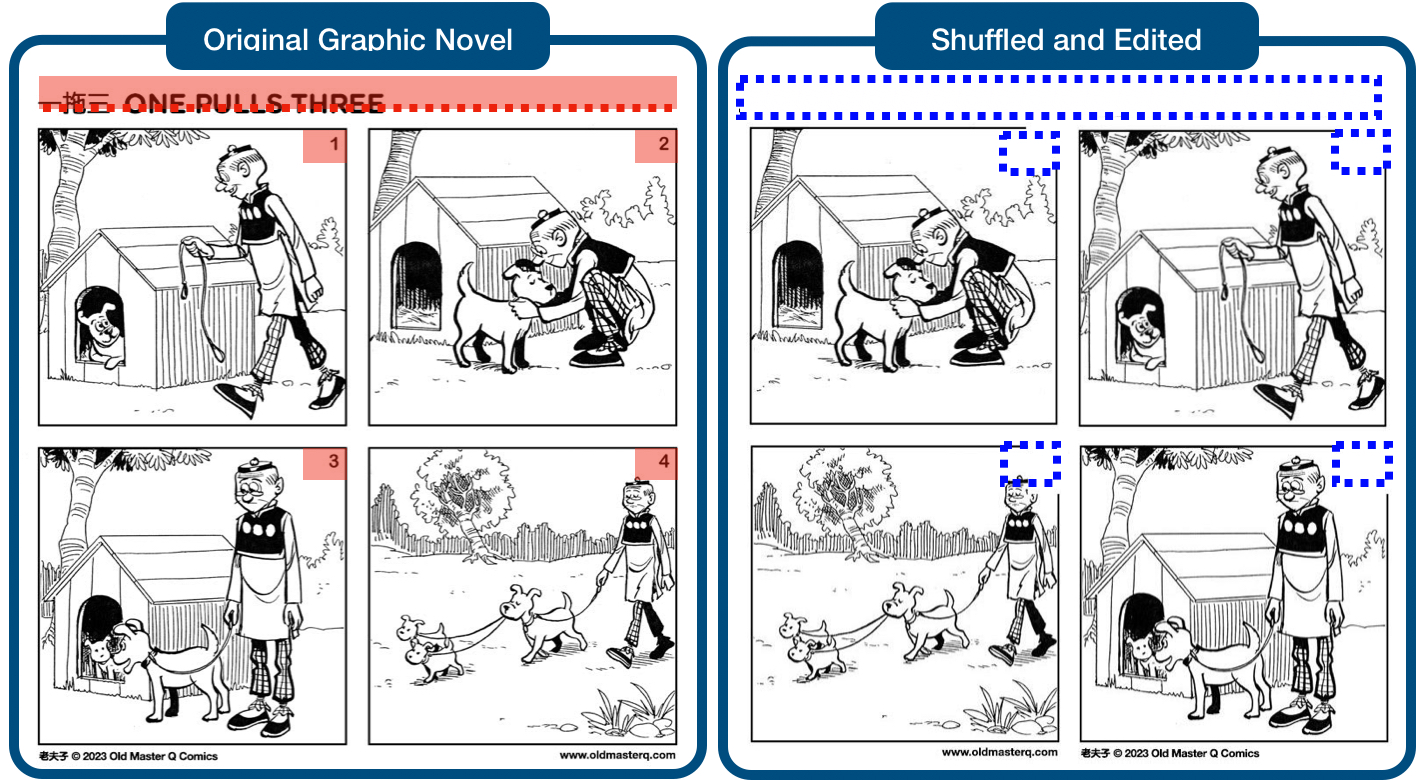}}
    \caption{
    We remove clue-containing sections marked by red boxes that help determine the correct storyline. These sections were removed as shown by the blue dotted line boxes in the 'Shuffled and Edited' version.
    }
    \label{grapchic_f1old}
\end{figure}

\begin{figure*}[ht]
    \centering
    \includegraphics[width=\columnwidth]{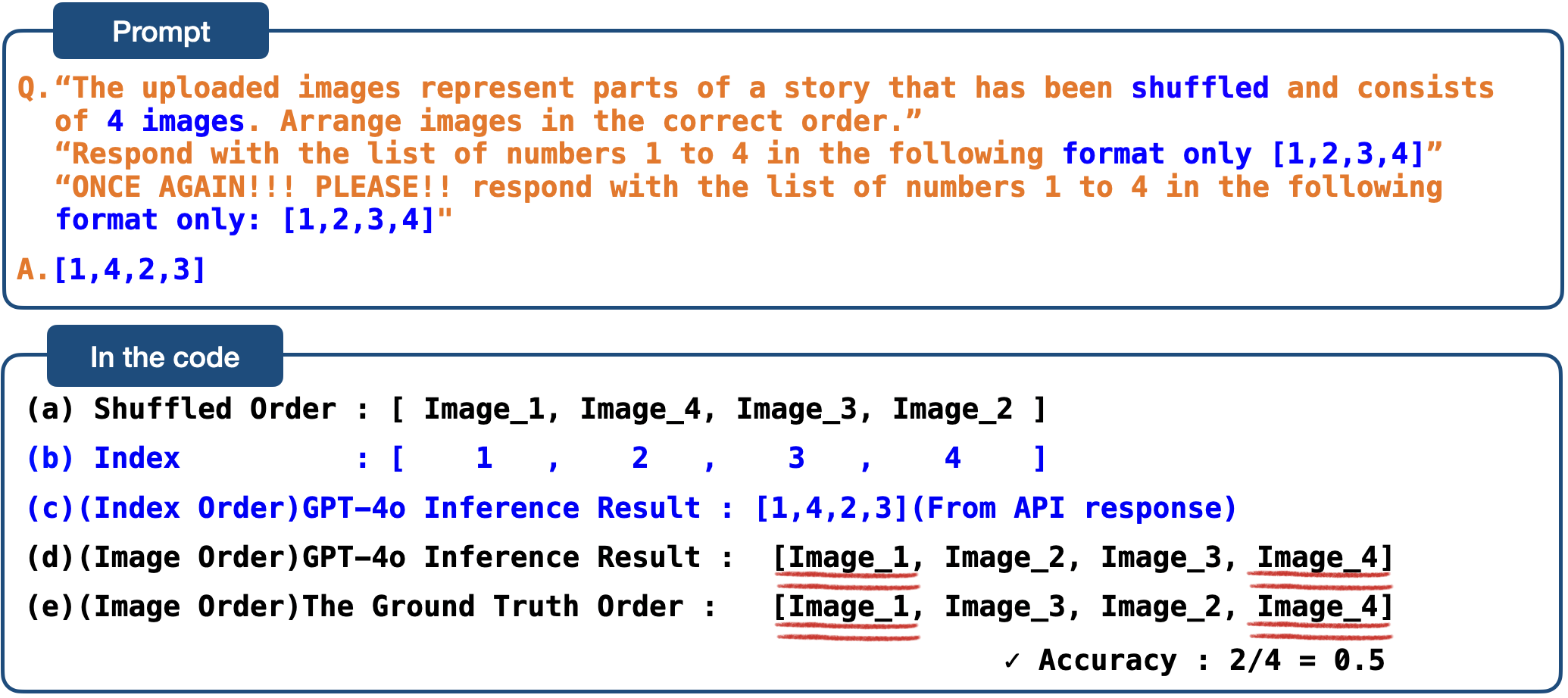} 
    \caption{Description of the random shuffle experiment process: In the 'Prompt', all essential information is provided, including the fact that all images are shuffled, that the four images are parts of a story, and the response format. The 'Code' section illustrates the task sequence from (a) to (e). (a) shows the shuffled input image order, (b) is the index of the input image order, (c) is GPT-4o's response which is the inferred result, (d) is the transformation from index order to image order, and (e) is the ground truth order used to calculate accuracy.}
    \label{graphic_f2} 
\end{figure*}

We utilized graphic novels, which are rich in content and often depict exceptional cases, to test the FMs' understanding. The experiment involves short story graphic novels: four-panel graphic novels with shuffled sequences, where the task for the FMs is to rearrange the panels into the correct order. We selected 'Old Master Q Comics' \citep{Wong1983} for this purpose, as these graphic novels revolve around comedy and typically have short storylines. These present vividly exaggerated storylines that are seldom encountered by FMs.\\
\textbf{Data Details:} We collected the graphic novels through web scraping and then segmented them panel by panel using automated Python scripts. We reviewed and excluded data entries that contained unevenly sized panels to maintain consistency in the dataset. This dataset allows us to evaluate the extent to which the FMs comprehend the storyline. To ensure an accurate assessment, we eliminate all clues that provide information about the storyline, including panel numbers and titles of the graphic novel as shown in Figure \ref{grapchic_f1old}.\\
\textbf{Experiments Details:} The API temperature setting is adjusted to 0.01 for Gemini-1.5-Pro and 0 for GPT-4 to ensure consistent results. To generate a concise answer, the model is instructed to output the response solely in the format [1,2,3,4], as shown in the blue text in Figure \ref{graphic_f2} ('Prompt'). We set the ground truth order as [1,3,2,4] to automate the task, given that the input images are shuffled, as shown in (e) in Figure \ref{graphic_f2} ('In the code'). This predetermined order allows us to verify whether FMs produces the correct sequence. Additionally, we demonstrate how the prompts were designed for each style in E.1. Table \ref{tab:prompts1}.
We design the random shuffle experiment as follow.\\
1. Inform the FMs that the uploaded images represent parts of a story that have been shuffled and consist of four images as shown in the blue letters in Figure \ref{graphic_f2} ('Prompt'). Instruct it to analyze all the images and deduce the correct sequence.\newline
2. Upload four images in a shuffled order, with each image assigned an ID number as shown in (a), (b) in Figure \ref{graphic_f2} ('In the code').\newline
3. The uploaded images are indexed, and the FMs infers the correct order, subsequently outputting the images in the proper indexed sequence as shown in (c) in Figure \ref{graphic_f2} ('In the code').\newline
4. Using code, the indexed sequence is transformed into a sequence of image ID numbers to obtain the image order predicted by the FMs as shown in (d) in Figure \ref{graphic_f2} ('In the code').\newline
5. Compare the predicted image order with the ground truth order to determine accuracy as shown in (e) in Figure \ref{graphic_f2} ('In the code').

\begin{table*}[t]\vspace{-1.0em}
\caption{
We tried many other version of CoT to enhance capability of GPT-4o on Graphic Novels dataset such as the prompt in this table.
}
\label{tab:prompts}\vspace{-1.0em}
\resizebox{\textwidth}{!}{%
\centering
\begin{tabular}{cl}
\toprule
{Graphic Novels}\\
\\  \midrule
{Example} & {Prompt}\\
\\  \midrule \cline{1-2}
CoT (Detailed Multi-Step Version)
& \begin{tabular}[c]{@{}l@{}}
\\
Input : Q. “The uploaded images represent parts of a story that has been shuffled and consists of 4 images."\\ "Arrange images in the correct order.”\\
IMPORTANT: Respond ONLY with the list of numbers 1 to 4 in this format: [1, 2, 3, 4].\\
\\
A. Let's think step by step.\\
1. Initial Observation: Look at the comic image for a moment. What stands out to you immediately?\\
2. Setting Description: Describe the setting. Where does the scene take place? Include details about the background and environment.\\
3. Character Identification: Who are the characters in the image? Describe their appearance and any notable features.\\
4. Actions and Interactions: What are the characters doing? Describe their actions and how they interact with each other.\\
5. Text Elements: What text elements are present? What are the characters saying or thinking, and how does this contribute to the scene?\\
6. Emotional Tone and Atmosphere: What is the emotional tone of the scene? Describe the mood and emotions conveyed by the characters and setting.\\
7. Context and Story Progression: What do you think happened before this scene, and what might happen next? How does this image fit into the larger story?\\
8. Summary and Interpretation: Summarize your description. What is the key aspect of this comic image, and what theme or message does it convey?\\
\\
By these logical steps, the correct order of the images is:\\
Output: A. \end{tabular}      \\\hline
\end{tabular}
}
\end{table*}
\subsection{Task Result}
 \label{appendix: A.2}
\begin{figure}[ht]\vspace{-0.5em}
    \centerline{\includegraphics[width=\columnwidth]{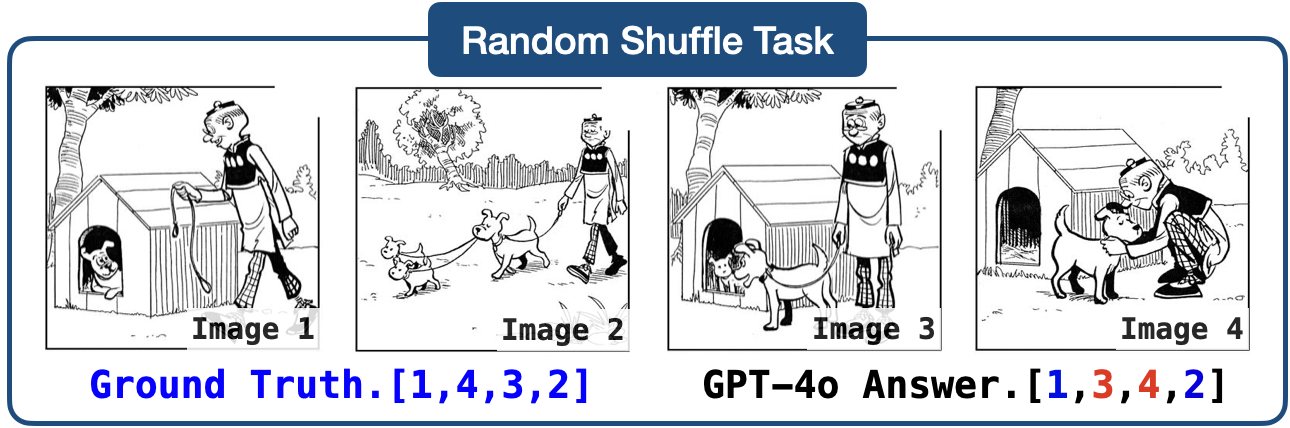}}
    \vspace{-0.5em} 
    \caption{Example of the random shuffle task. The original sequence is [1, 4, 3, 2], but GPT-4o produce an incorrect result.
    }
    \label{graphic_f1} 
\end{figure}

\begin{figure}[ht]
    \centering
    \includegraphics[width=\columnwidth]{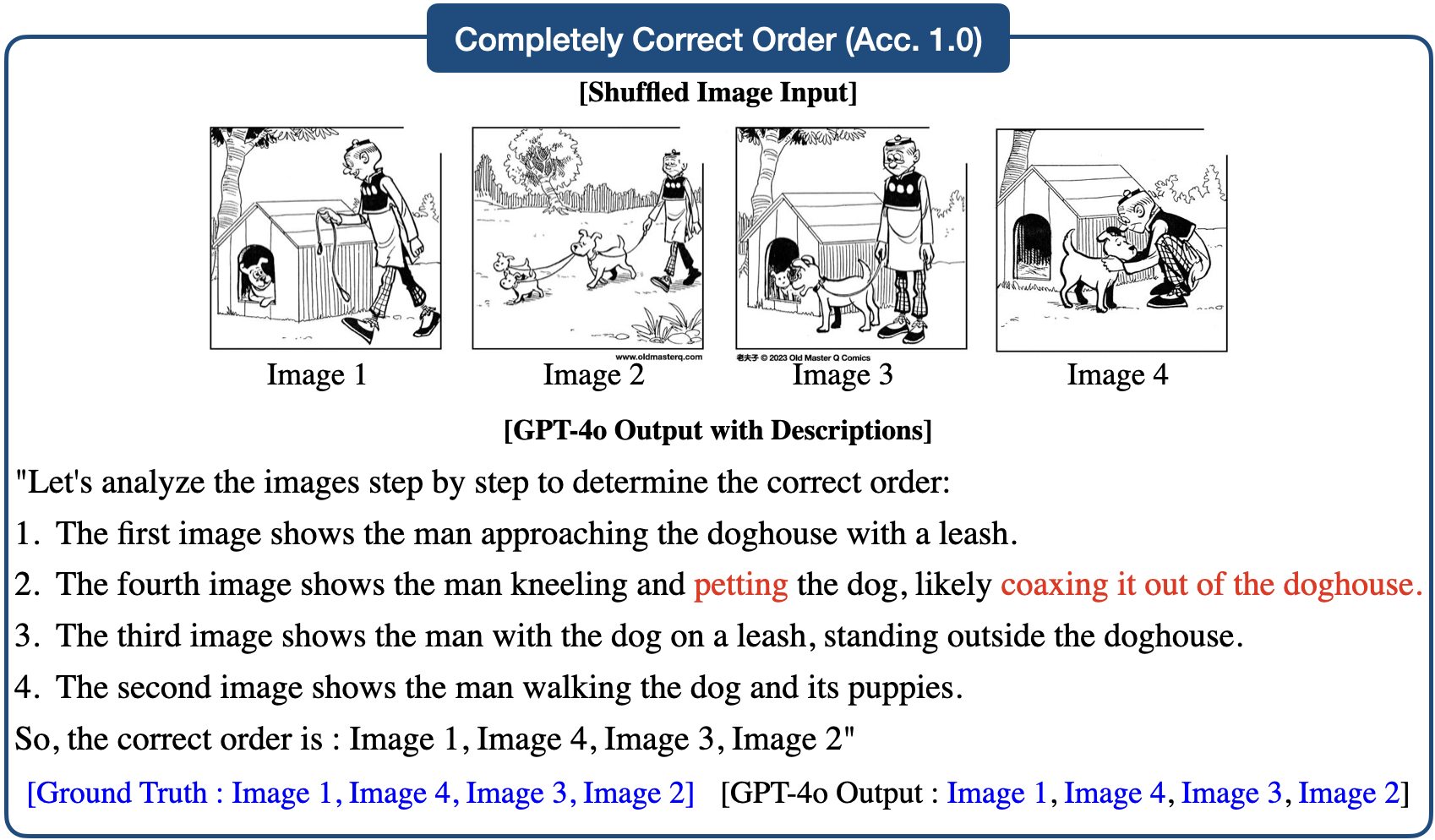}\vspace{-0.5em} 
    \caption{Correct Order Check: This example shows that while GPT-4o can correctly order the images, it sometimes lacks in scene description such as using mismatched verbs (highlighted in red).}
    \label{graphic_figure5} 
\end{figure}

\begin{figure}[ht]
    \centering
    \includegraphics[width=\columnwidth]{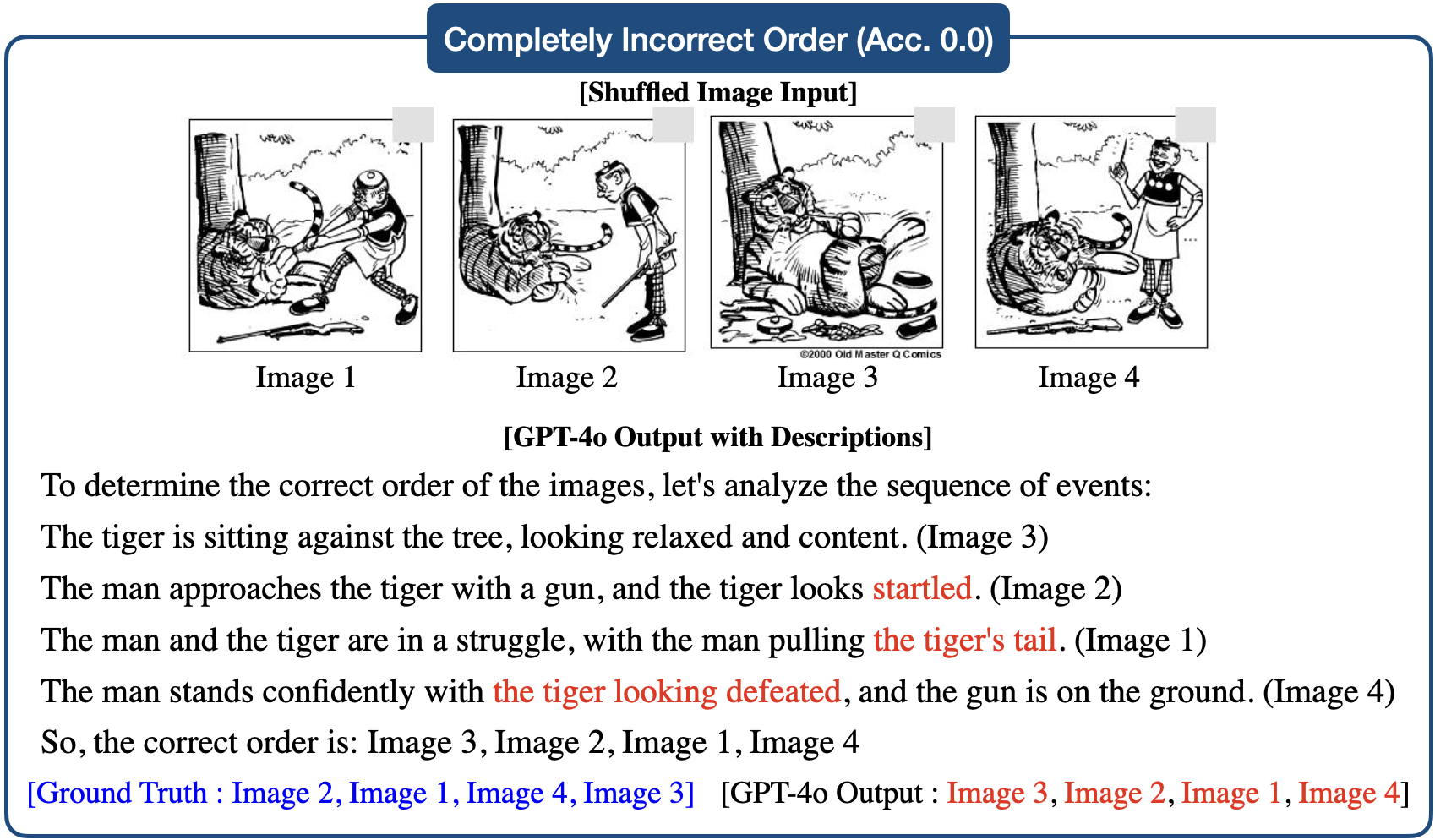}\vspace{-0.5em} 
    \caption{Incorrect Order Check: In three of four images, GPT-4o provided incorrect character descriptions and showed poor object recognition (highlighted in red).}
    \label{graphic_figure14} 
\end{figure}

We assessed the multimodal causal reasoning abilities of FMs through a Random Shuffle task. We hypothesize that if FMs can comprehend the story lines through causal reasoning, it is likely to be able to infer the correct sequence of panels when presented with a randomly shuffled input. Based on this hypothesis, we designed the random shuffle task as shown in Figure \ref{graphic_f1}. The highest performance was observed in the CoT+Few-Shot condition, followed by CoT and then Zero-Shot. Interestingly, the Zero-Shot performance exceeded expectations, displaying an accuracy that was not markedly lower than the other prompting styles. During the CoT style prompt experiments, we conducted various tests ranging from the very simple 'Let's think step by step' to more detailed descriptions of the reasoning sequence as shown in Table \ref{tab:prompts}. Interestingly, the simplest 'Let's think step by step' prompt yielded the best performance. There was some variation depending on whether 'Let's think step by step' was prompted before or after the task images. In the case of CoT+Few-Shot, the number of Few-Shot examples impacted performance; with only one example, there was no difference compared to CoT, but increasing the examples to three resulted in a noticeable performance improvement.
 \\
\textbf{When the inferred order is completely correct:} FMs occasionally makes mistakes in scene descriptions, even when it derives correct answers. For example, in Figure \ref{graphic_figure5}, GPT-4o describes a man as 'kneeling and petting the dog, coaxing it out of the doghouse,' whereas the actual scene is 'squatting in front of the doghouse, putting a leash on the dog.'\\
\textbf{When the inferred order is completely incorrect:} FMs sometimes misidentify objects or misunderstand emotions. For instance, GPT-4o describes a man pulling a tiger's tail instead of removing an arrow from its paw, refer to image 2 of Figure \ref{graphic_figure14}.
\newline

\section{Calligraphy}
\subsection{Task Details}
 \label{appendix: B.1}
\begin{figure}[h]
    \centerline{\includegraphics[width=\columnwidth]{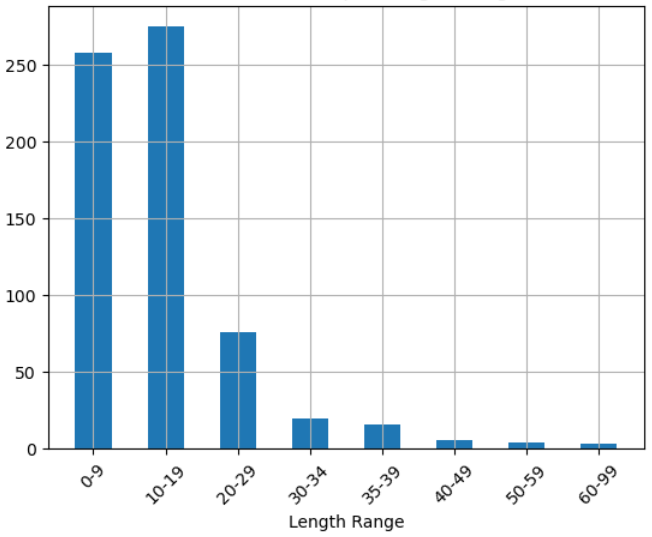}}
    \vspace{-0.5em} 
    \caption{Length plot of Korean calligraphy images. We determined that images with over 35 characters presented considerable visual recognition difficulties, even for humans, prompting their exclusion from our evaluation.}
    \label{fig:calligrpahy len}
\end{figure}

\begin{figure}[ht]\vspace{-1.0em}
    \centerline{\includegraphics[width=\columnwidth]{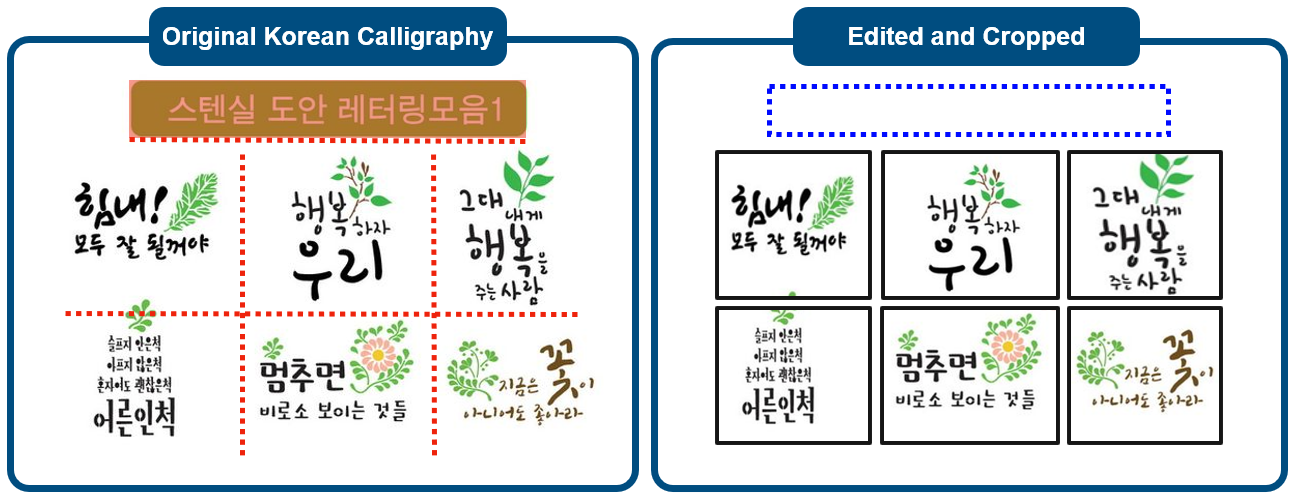}}
    \vspace{-0.5em} 
    \caption{Example of preprocessed Korean calligraphy. We removed typographic elements unrelated to the calligraphy and automatically cropped overlapping sections using bounding boxes detected by the OCR API.}
    \label{fig: calligraphy pre} 
\end{figure}

\textbf{Data Details:} We preprocessed the dataset according to three rules. First, we deleted images if their resolution was too low or if they contained too many letters that even a human could not recognize. We set the threshold at 35 characters, as shown in Figure \ref{fig:calligrpahy len}, where 35 is an irregularly large number in the dataset. We observed that images with more than 35 characters are visually challenging for humans to recognize, so we excluded such images from evaluation. Second, we separated overlapping calligraphy in an image by applying bounding boxes provided by the OCR API. Third, we cropped out typographic elements such as signs and watermarks that were deemed irrelevant to the calligraphy. An example of the preprocessed Korean calligraphy is shown in Figure \ref{fig: calligraphy pre}.\\
\newline
\textbf{Experiments Details:} The API temperature setting is adjusted to 0.01 for Gemini-1.5-Pro and 0 for GPT-4 to ensure consistent results. Before word-level evaluation, we removed punctuation and special symbols from FM predictions and replaced '\textbackslash n' with ' ' due to ambiguous line breaks in the calligraphy. We used Word-level Accuracy, CER, and WER, which are representative OCR metrics.

\subsection{Task Result}
 \label{appendix: B.2}
\begin{figure}[ht]
    \centerline{\includegraphics[width=\columnwidth]{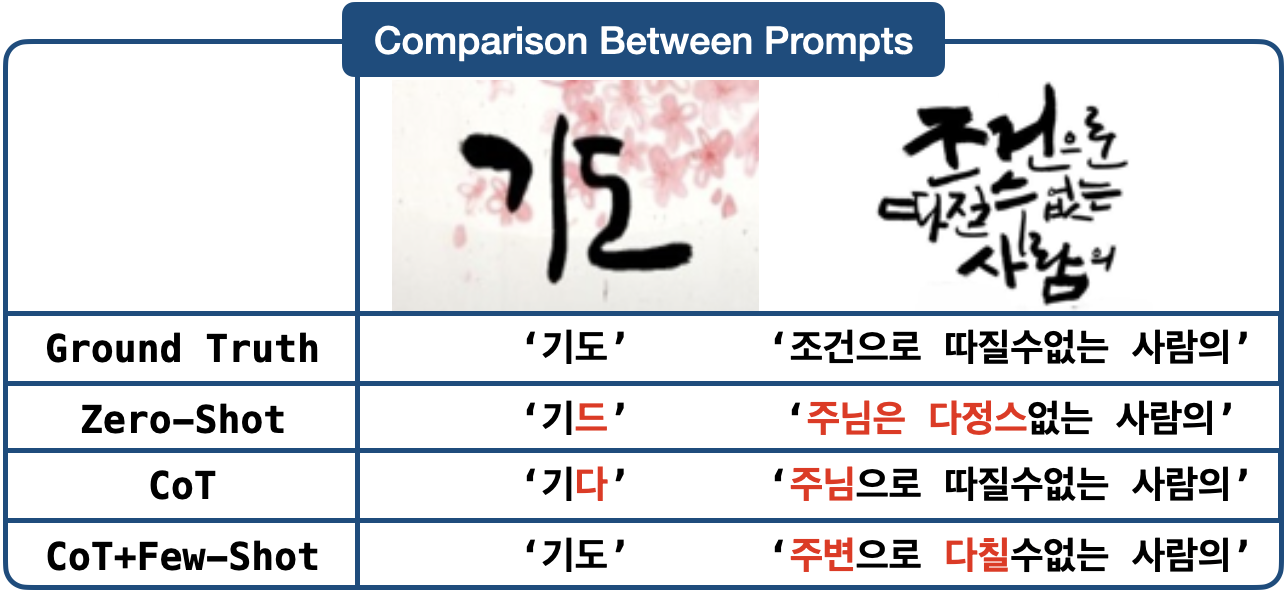}}
    \caption{Examples of comparisons of OCR task results between prompts on Korean calligraphy data.}
    \label{fig: calli compar promp} 
\end{figure}

 The artistic nature of calligraphy sometimes leads to unconventional representations in the dataset, such as abbreviating 'spring day' to 'spring d.' In these cases, FMs tend to process 'd' as a separate element rather than part of the word, recognizing only 'spring.' This tendency was more pronounced in the CoT and CoT+Few-Shot prompts compared to Zero-Shot. In the Zero-Shot scenario, the OCR task tends to prioritize the visual recognition of individual words over the holistic meaning conveyed by the calligraphy, resulting in a higher frequency of typographical errors. Conversely, the CoT and CoT+Few-Shot approaches first interpret the overall meaning and then perform OCR based on contextually relevant words. Consequently, even when the output deviates from the ground truth, it tends to generate semantically similar words or words that are more contextually fitting than the ground truth. As illustrated in Figure \ref{fig: calli compar promp}, the first calligraphy example signifies 'pray,' with the ground truth being '기도.' In the Zero-Shot scenario, GPT-4o recognizes it as '기드,' which bears a close visual resemblance but lacks semantic meaning. The CoT approach interprets it as '기다,' which, although not aligning with the ground truth, at least carries the meaning 'to crawl.' Notably, the CoT+Few-Shot approach accurately identifies it as '기도,' precisely matching the ground truth.
\section{Onion, Not The Onion}
\subsection{Task Details}
 \label{appendix: C.1}
\begin{figure}[ht]\vspace{-0.5em}
    \centerline{\includegraphics[width=\columnwidth]{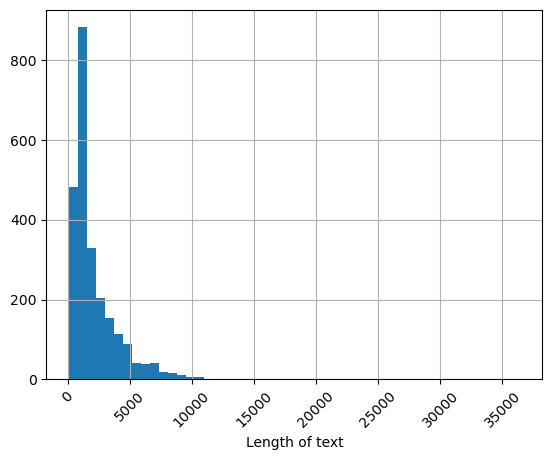}}
    \vspace{-0.5em} 
    \caption{Length plot of the preprocessed Onion and Not the Onion news data.}
    \label{fig: fakenews length plot}\vspace{-0.5em}
\end{figure}

\textbf{Data Details:} We performed web scraping on The Onion website and Reddit's Not The Onion section. Following data collection, we implemented an additional filtering process using Python scripts to enhance the dataset's sophistication. Specifically, we automated the removal of instances where no content was collected, where content was duplicated, and where advertisements were included. During preprocessing, we encountered valid data with varying lengths, both long and short, that were indeed written by humans. These instances represent qualitative news articles, so we chose not to remove them to preserve the dataset's integrity. As a result, the mean and median text lengths are 2243 and 1433, respectively, leading to a left-skewed distribution. A histogram illustrating text lengths and category-specific statistics is presented in Figure \ref{fig: fakenews length plot}. Through this process, we ensured that only the title and content of the original news articles influenced the FMs' judgment during fake news detection. This approach provided a reliable dataset, allowing us to evaluate the impact of textual data alone in fake news detection research.\\
\textbf{Experiments Details:} Recent studies have demonstrated that proper prompting can enhance the performance of FMs \citep{kojima2022large}. In this study, The default prompt simply asked the model to distinguish between fake news and real news. In contrast, the CoT prompts instructed the model to go through a step-by-step process of thinking to determine fake news \citep{wei2022chain}. In this methodology, the model is instructed to take specific thought steps. Finally, we measured the performance of the model for the Few-shot and CoT prompts by providing examples of fake news and real news, as well as illustrating the judgment process. Through these comparisons, we evaluated the impact of various prompting methods on the model's ability to recognize fake news. The detailed prompts are provided in Table \ref{tab:fakenews_prompts}. By distinguishing between fake news and real news, we contribute to preventing social disruption and maintaining the credibility of information.
\vspace{-0.5em}

\begin{table}[ht]
\caption{Comparison of performance metrics between Gemini-1.5-Pro and GPT-4o across different settings.}
\label{tab:performance-comparison}\vspace{-0.5em}
\resizebox{\columnwidth}{!}{
\centering
\begin{tabular}{ccccccc}
\hline
\multicolumn{2}{c}{\textbf{Metric}} & \textbf{Model}  & \textbf{Zero-Shot} & \textbf{CoT}    & \textbf{CoT+Few-Shot} \\ \hline
\multirow{2}{*}{\textbf{Acc.}}  &        & \textbf{Gemini-1.5-Pro} & 83.97     & 87.81 & 91.91        \\ \cline{3-6} 
                                &        & \textbf{GPT-4o}         & 80.70     & 89.88 & 94.74        \\ \hline
\multirow{3}{*}{\textbf{Onion}} & \textbf{Precision} & \textbf{Gemini-1.5-Pro} & 84.12     & 84.84 & 88.07        \\ \cline{3-6} 
                                &        & \textbf{GPT-4o}         & 78.70     & 86.14 & 92.49        \\ \cline{2-6} 
                                & \textbf{Recall}    & \textbf{Gemini-1.5-Pro} & 87.27     & 94.43 & 98.42        \\ \cline{3-6} 
                                &        & \textbf{GPT-4o}         & 85.19     & 95.52 & 97.60        \\ \cline{2-6} 
                                & \textbf{F1-score}  & \textbf{Gemini-1.5-Pro} & 85.67     & 89.38 & 92.96        \\ \cline{3-6} 
                                &        & \textbf{GPT-4o}         & 81.81     & 90.58 & 94.97        \\ \hline
\multirow{3}{*}{\textbf{Not The Onion}} & \textbf{Precision} & \textbf{Gemini-1.5-Pro} & 83.78     & 92.35 & 97.82        \\ \cline{3-6} 
                                        &        & \textbf{GPT-4o}         & 83.17     & 94.75 & 97.35        \\ \cline{2-6} 
                                        & \textbf{Recall}    & \textbf{Gemini-1.5-Pro} & 79.96     & 79.94 & 84.18        \\ \cline{3-6} 
                                        &        & \textbf{GPT-4o}         & 76.04     & 84.03 & 91.76        \\ \cline{2-6} 
                                        & \textbf{F1-score}  & \textbf{Gemini-1.5-Pro} & 81.82     & 85.70 & 90.49        \\ \cline{3-6} 
                                        &        & \textbf{GPT-4o}         & 79.44     & 89.07 & 94.48        \\ \hline
\end{tabular}
}\vspace{-0.5em}
\end{table}

\subsection{Task Result}
 \label{appendix: C.2}
\begin{figure}[h]
    \centerline{\includegraphics[width=\columnwidth]{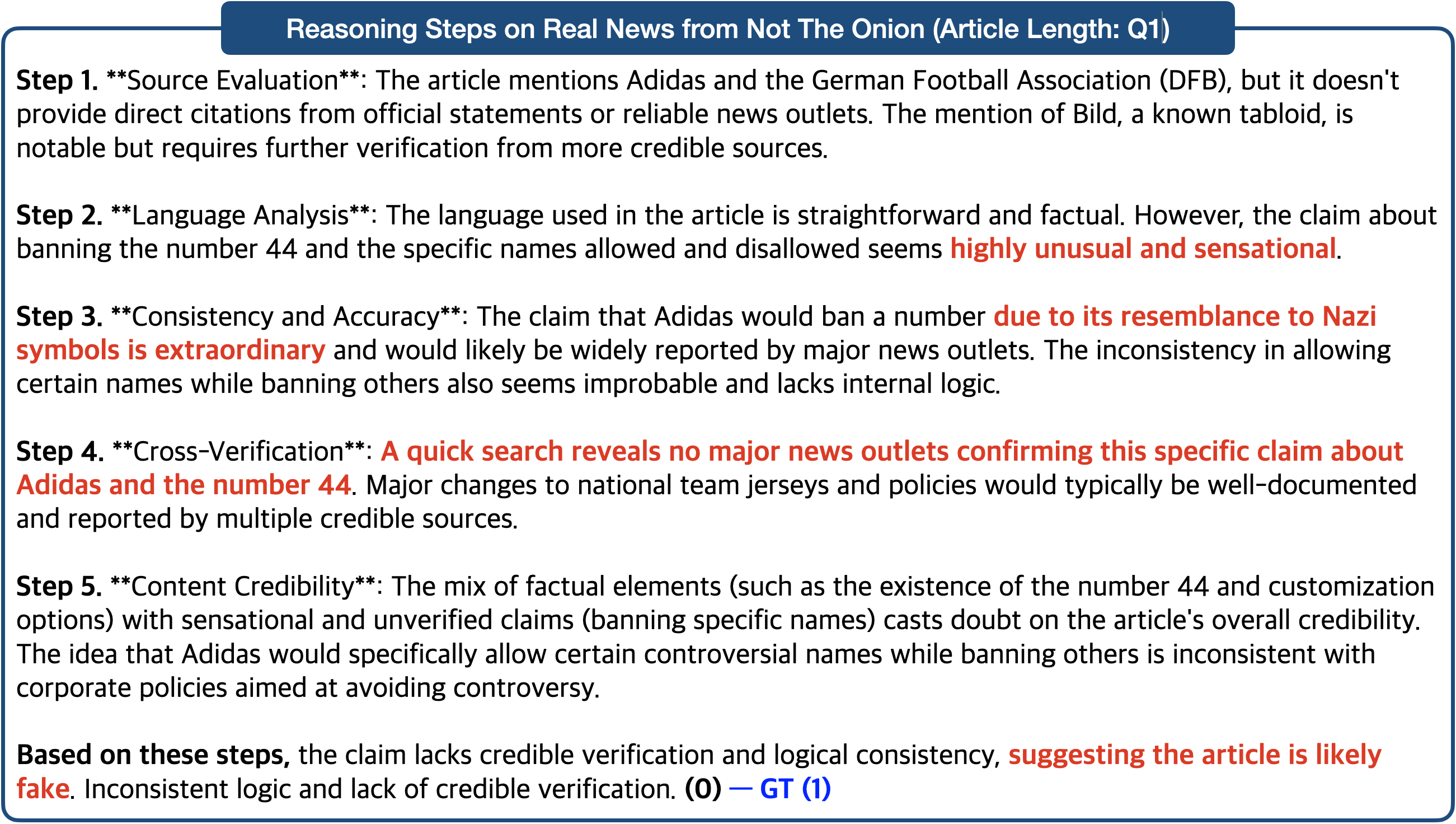}}
    \caption{An example where GPT-4o, despite following appropriate reasoning steps, produces an incorrect reasoning outcome due to an exceptional case(highlighted in red).}
    \label{fig: false negative} 
\end{figure}

Overall, FMs exhibit high performance on Onion, Not The Onion dataset as shown in Table \ref{tab:performance-comparison}, but we observed a reduction in performance with relatively short articles. As shown in Table \ref{tab:onion-result}, accuracy differences based on article length reveal that accuracy generally improves as article length increases. In contrast, the Onion group, predominantly consisting of fake news articles, typically features shorter articles and maintains consistently high accuracy across the dataset. This pattern suggests that FMs may have a tendency to classify shorter articles as fake news, highlighting the greater challenge posed by Not The Onion in fake news classification. Furthermore, we delve deeper into the rationale behind FMs's decision-making process, particularly when encountering relatively short articles, to better understand the circumstances under which FMs arrives at incorrect conclusions and whether it follows appropriate steps in such cases. In this approach, we observe that FMs generally takes appropriate steps, many of which are plausible. However, it is notable that FMs encounters difficulties with exceptional cases, as highlighted in Figure \ref{fig: false negative}(marked in red). The article depicted in this figure includes several extraordinary claims, such as "Adidas urgently recalled the German national team jersey featuring the number 44 due to its resemblance to symbols used by the German SS division". 
To verify these claims, GPT-4o undergoes a validation process spanning from the second to the fourth step. Despite employing a search function in the fourth step, it fails to accurately determine the veracity of the article.
Overall, to identify fake news, GPT-4o needs accurate causal reasoning to classify instances within an article. This makes the Onion, Not the Onion dataset a splendid benchmark for verifying their reasoning capabilities.
\begin{figure}[ht]
    \centerline{\includegraphics[width=\columnwidth]{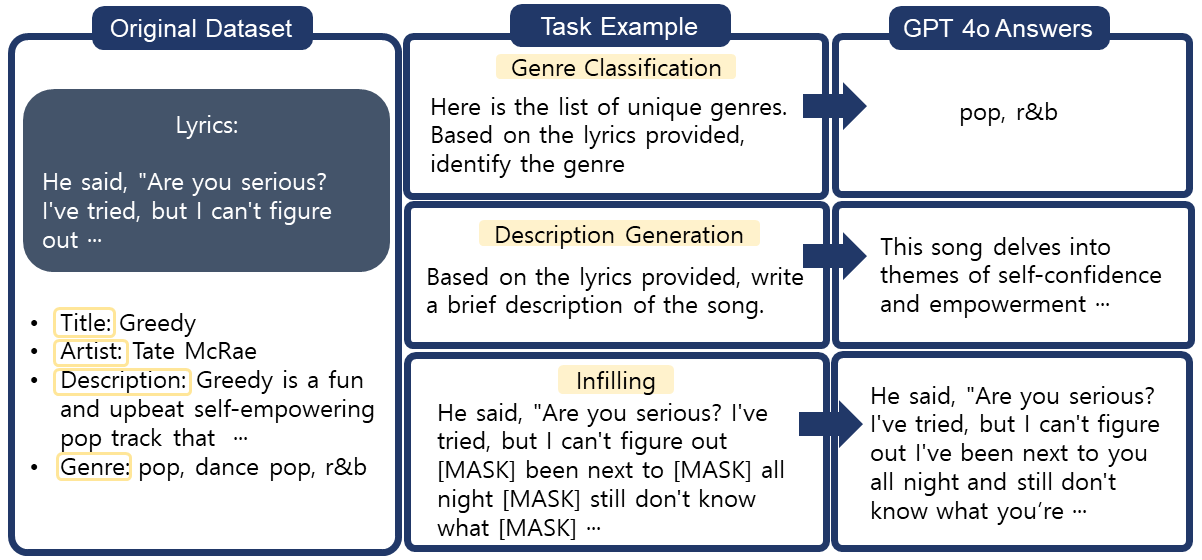}}
    \vspace{-0.5em} 
    \caption{Overview of the lyrics dataset: an example of three different tasks and GPT-4o's responses.}
    \label{fig:lyrics example}
    \vspace{-0.5em} 
\end{figure}\vspace{-0.5em}
\section{Lyrics}
\subsection{Task Details}
 \label{appendix: D.1}
\textbf{Data Details:} Although lyrics often contain poetic licenses and uncommon expressions such as metaphors, song lyrics still allow for meaningful inference as one of the main literary genres. To evaluate the robustness of reasoning capabilities in FMs when dealing with exceptional data like lyrics, we constructed a dataset using song lyrics. We assess FMs’ comprehension of song lyrics through three tasks: genre detection, song description generation, and infilling as shown in Figure \ref{fig:lyrics example}. For the infilling task, we used a pre-trained BERT model to anticipate the masked parts and removed non-exceptional data. Entries with BERT scores exceeding a 0.9 threshold were excluded, as high semantic similarity indicated non-exceptional content. When collecting the dataset, we divided it into two parts: 'yearly' and 'weekly.' The yearly dataset comprises data from before the FMs cut-off date (before the end of 2023), while the weekly dataset includes data from after the cut-off date (after the end of 2023). For the English dataset, after collecting the title and artist of each song, we removed duplicate entries—only removing a song if both the title and artist were identical, as different songs can share the same title. We then generated links to the Genius site to obtain the lyrics and descriptions of the songs. This process involved removing strings following 'featuring' and modifying characters such as brackets and Latin alphabets. If it was impossible to retrieve any of the descriptions, genre, or lyrics due to link generation errors or unavailability on the site, we excluded the song. Additionally, songs with non-English lyrics were also removed. To ensure that the weekly dataset contained only data that the FMs had not previously encountered, any song appearing in both the weekly and yearly data was excluded from the weekly dataset. For the genre detection task in English, we streamlined the genre list by removing infrequent genres. After consolidating all genre lists, we excluded genres with fewer than 10 occurrences, resulting in a final list of 58 unique genres and a dataset of 1,811 songs. A similar process was applied to both the English and Korean datasets. However, for the Korean dataset, non-Korean lyrics were not removed due to their high frequency, and genre cleaning was not performed because the dataset contained fewer genre categories. Notably, no songs were excluded during the crawling of lyrics, descriptions, or genres in the Korean dataset, as all song information was sourced from Melon, unlike the English dataset, which compiled data from multiple sites. The specific number of remaining data at each step is summarized in Table \ref{tab:lyrics preprocessing steps}.
\begin{table}[ht]
\caption{During the collection of song data, various criteria were used to remove certain songs, as detailed in the first column of the table. Numbers in each blocks denotes the number of remaining data after each step. X indicates that the dataset did not go through that step.}
\label{tab:lyrics preprocessing steps}
\resizebox{\columnwidth}{!}{%
\begin{tabular}{lcccc}
\hline
\multicolumn{1}{c}{\multirow{2}{*}{\textbf{}}} & \multicolumn{2}{c}{\textbf{English}} & \multicolumn{2}{c}{\textbf{Korean}} \\ \cline{2-5} 
\multicolumn{1}{c}{} & \textbf{Before Cut-Off} & \textbf{After Cut-Off} & \textbf{Before Cut-Off} & \textbf{After Cut-Off} \\ \hline
\textbf{Total} & 3400 & 1700 & 3400 & 1700 \\ \hline
\textbf{Delete duplicate songs} & 3112 & 353 & 2187 & 304 \\ \hline
\textbf{\begin{tabular}[c]{@{}l@{}}Lyrics and Description\\ crawling\end{tabular}} & 2435 & 246 & 2187 & 304 \\ \hline
\textbf{Genre crawling} & 1828 & 139 & 2187 & 304 \\ \hline
\textbf{Remove Multilingual} & 1803 & 131 & X & X \\ \hline
\textbf{\begin{tabular}[c]{@{}l@{}}Remove duplicate between \\ yearly and weekly\end{tabular}} & X & 121 & X & 176 \\ \hline
\textbf{Cleaning Genre} & 1703 & 108 & X & X \\ \hline
\textbf{Final} & 1703 & 108 & 2187 & 176 \\ \hline
\end{tabular}%
}
\end{table}
\begin{table}[ht]
\caption{Evaluation metric of each task using lyrics. Empty block denotes that we did not used the data for the corresponding task.} 
\label{tab:lyrics task evaluation metrics}
\resizebox{\columnwidth}{!}{%
\begin{tabular}{cclll}
\hline
\multicolumn{2}{c}{} & \multicolumn{1}{c}{\textbf{\begin{tabular}[c]{@{}c@{}}Genre\\ Classification\end{tabular}}} & \multicolumn{1}{c}{\textbf{\begin{tabular}[c]{@{}c@{}}Description\\ Generation\end{tabular}}} & \multicolumn{1}{c}{\textbf{\begin{tabular}[c]{@{}c@{}}Lyrics\\ Infilling\end{tabular}}} \\ \hline
\multirow{2}{*}{\textbf{Korean}} & \textbf{Before Cut-Off} & \begin{tabular}[c]{@{}l@{}}- Overlap Ratio\\ - Exact Match\end{tabular} &  &  \\ \cline{2-5} 
 & \textbf{After Cut-Off} & \begin{tabular}[c]{@{}l@{}}- Overlap Ratio\\ - Exact Match\end{tabular} &  & \begin{tabular}[c]{@{}l@{}}- ROUGE\\ - BERT Score\end{tabular} \\ \hline
\multirow{2}{*}{\textbf{English}} & \textbf{Before Cut-Off} & \begin{tabular}[c]{@{}l@{}}- Overlap Ratio\\ - Exact Match\end{tabular} & \begin{tabular}[c]{@{}l@{}}- ROUGE\\ - BERT Score\end{tabular} &  \\ \cline{2-5} 
 & \textbf{After Cut-Off} & \begin{tabular}[c]{@{}l@{}}- Overlap Ratio\\ - Exact Match\end{tabular} & \begin{tabular}[c]{@{}l@{}}- ROUGE\\ - BERT Score\end{tabular} & \begin{tabular}[c]{@{}l@{}}- ROUGE\\ - BERT Score\end{tabular} \\ \hline
\end{tabular}%
}
\end{table}\\
\textbf{Experiments Details:} We employed several metrics for precise testing, including BERT Score and ROUGE, which are well-known, as well as Exact Match and Overlap Ratio, specifically utilized for this dataset as shown in Table \ref{tab:lyrics task evaluation metrics}. An Exact Match score assigns a value of 1 if the predicted genre matches the original genre. The Overlap Ratio measures similarity based on shared elements. The F1 score reflects the extent of overlap between the generated answer and the ground truth. Recall scores were used to confirm whether the original lyrics were present within the words generated by the FMs. The model is instructed to generate answers in specific formats: for the Genre classification task, "Genre: {the output}"; for the song description generation task, "Description: {the output}"; and for the infilling task, FMs should provide the complete lyrics, including the predicted masked part. Additional details about the prompts are in Appendix \ref{appendix: E.4}\\
\textbf{Genre classification:} We design the genre classification task as follow:\newline
1. A unique genre list was created by concatenating all possible genres and removing entries with fewer than 10 occurrences. This reduced the size of the genre lists and removed datasets with no genres.\newline
2. We conducted separate experiments on the Before Cut-Off dataset, which includes data from 1990 to 2023, and the After Cut-Off dataset, covering January to April 2024. This was done to determine if there is a performance difference between the periods that FMs has been trained on and those it has not.\newline
3. FMs was then asked to select the most likely genre(s) based on the provided lyrics.\newline
4. For the zero-shot approach, FMs generated the output directly. For the CoT and CoT+Few-shot prompts, FMs was instructed to think in alignment with the lyrics.\\
\textbf{Description generation:} 
We design the description generation task as follow:\newline
1. FMs was asked to generate a song description based on the provided lyrics.\newline
2. We conducted separate experiments on the seen dataset, which includes data from 1990 to 2023, and the unseen dataset, covering January to April 2024. This was done to determine if there is a performance difference between the periods that FMs has been trained on and those it has not.\newline
3. Since many ground truth song descriptions included additional information about the song (e.g., interviews, messages to fans, or musical features), for the CoT and CoT+Few-shot prompts, we included instructions for FMs to add possible artist names, title names, and musical features.\\
\textbf{Infilling:}
We design the infilling task as follow:\newline
1. For the English seen and unseen datasets, masking was performed based on both word and token criteria to determine which masking technique would be more challenging.\newline
2. Using BERT, we compared the two masking methods: the average score for word-based masking was lower, so we decided to use the word-based masking dataset\newline
3. The Korean unseen dataset was also masked based on words, without the process described in step 1.
4. The infilling task was performed on the Korean and English datasets using BERT.\newline
5. The results from step 4 were evaluated using the BERT score. Data with scores exceeding 0.9 were removed.\newline
6. After step 5, the remaining data was used to perform the infilling task with FMs. Due to FMs's safety issues, only the After Cut-Off dataset was used.
\subsection{Task Result}
 \label{appendix: D.2}
\begin{figure}[t]
    \centerline{\includegraphics[width=\columnwidth]{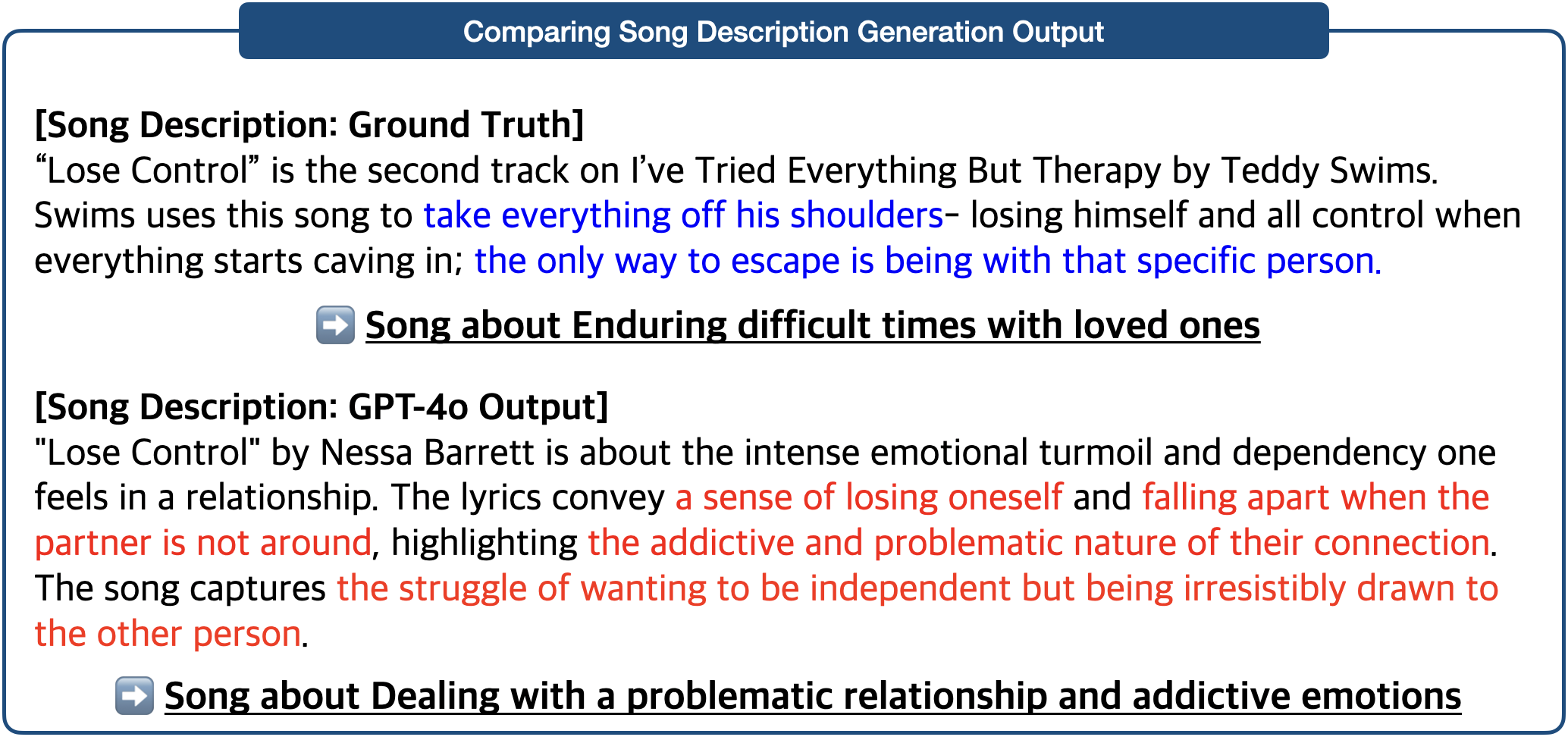}}
    \caption{In the description generation task, It is evident that FMs does not accurately comprehend song lyrics. In the example, unlike the ground truth, which refers to 'enduring difficult times with loved ones,' GPT-4o generated content describing 'dealing with a problematic relationship and addictive emotions.'}
    \label{fig: description example} 
\end{figure}

In 2.Experiments and Results, we discussed the infilling task. Here, we focus on the genre classification and song description generation tasks.\\
\textbf{Genre Classification:} In the genre classification task, the difference in the number of unique genres between the English and Korean datasets influenced the results: 11 genres in Korean and 58 in English. This made the task more challenging for the English dataset, leading to FMs struggling more with the English data than the Korean data, as shown in Table \ref{tab:lyrics Result genre}.
\begin{table}[ht]
\caption{Results of the genre classification task, GPT-4o generally outperforms Gemini-1.5-Pro across the entire dataset. Interestingly, after the cut-off, both baseline models showed better performance in Korean than in English.}
\label{tab:lyrics Result genre}\vspace{-0.5em}
\resizebox{\columnwidth}{!}{%
\begin{tabular}{lcccccc}
\hline
 &  &  & \textbf{Model} & \textbf{Zero-Shot} & \textbf{CoT} & \textbf{CoT+Few-Shot} \\ \hline
\multirow{8}{*}{\textbf{English}} & \multirow{4}{*}{\textbf{Before Cut-Off}} & \multirow{2}{*}{\textbf{Overlap Ratio}} & \textbf{Gemini-1.5-Pro} & 0.214 & 0.218 & 0.306 \\ \cline{4-7} 
 &  &  & \textbf{GPT-4o} & 0.594 & 0.610 & 0.620 \\ \cline{4-7} 
 &  & \multirow{2}{*}{\textbf{Exact Match}} & \textbf{Gemini-1.5-Pro} & 0.306 & 0.316 & 0.434 \\ \cline{4-7} 
 &  &  & \textbf{GPT-4o} & 0.758 & 0.774 & 0.781 \\ \cline{2-7} 
 & \multirow{4}{*}{\textbf{After Cut-Off}} & \multirow{2}{*}{\textbf{Overlap Ratio}} & \textbf{Gemini-1.5-Pro} & 0.405 & 0.429 & 0.550 \\ \cline{4-7} 
 &  &  & \textbf{GPT-4o} & 0.474 & 0.497 & 0.509 \\ \cline{4-7} 
 &  & \multirow{2}{*}{\textbf{Exact Match}} & \textbf{Gemini-1.5-Pro} & 0.486 & 0.514 & 0.657 \\ \cline{4-7} 
 &  &  & \textbf{GPT-4o} & 0.671 & 0.671 & 0.677 \\ \hline
\multirow{8}{*}{\textbf{Korean}} & \multirow{4}{*}{\textbf{Before Cut-Off}} & \multirow{2}{*}{\textbf{Overlap Ratio}} & \textbf{Gemini-1.5-Pro} & 0.619 & 0.581 & 0.609 \\ \cline{4-7} 
 &  &  & \textbf{GPT-4o} & 0.642 & 0.665 & 0.733 \\ \cline{4-7} 
 &  & \multirow{2}{*}{\textbf{Exact Match}} & \textbf{Gemini-1.5-Pro} & 0.652 & 0.615 & 0.642 \\ \cline{4-7} 
 &  &  & \textbf{GPT-4o} & 0.676 & 0.698 & 0.752 \\ \cline{2-7} 
 & \multirow{4}{*}{\textbf{After Cut-Off}} & \multirow{2}{*}{\textbf{Overlap Ratio}} & \textbf{Gemini-1.5-Pro} & 0.503 & 0.665 & 0.673 \\ \cline{4-7} 
 &  &  & \textbf{GPT-4o} & 0.668 & 0.690 & 0.750 \\ \cline{4-7} 
 &  & \multirow{2}{*}{\textbf{Exact Match}} & \textbf{Gemini-1.5-Pro} & 0.538 & 0.710 & 0.713 \\ \cline{4-7} 
 &  &  & \textbf{GPT-4o} & 0.710 & 0.733 & 0.776 \\ \hline
\end{tabular}%
}
\end{table}\\
\textbf{Description Generation:} In the description generation task, the overall scores are poor, indicating that FMs struggle to accurately understand the meaning of song lyrics as shown in Table \ref{tab:lyrics Result English description}. As illustrated in Figure \ref{fig: description example}, the song discusses 'enduring difficult times with loved ones,' while GPT-4o describes it as 'dealing with a problematic relationship and addictive emotions.
\begin{table}[ht]\vspace{-0.5em}
\caption{Description generation task for English songs. The low overall score shows FMs wrestle with understanding the meaning of lyrics.}
\label{tab:lyrics Result English description}\vspace{-0.5em}
\resizebox{\columnwidth}{!}{%
\begin{tabular}{cccccc}
\hline
 &  &  & \textbf{Zero-Shot} & \textbf{CoT} & \textbf{CoT+Few-Shot} \\ \hline
\multirow{18}{*}{\textbf{Before Cut-Off}} & \multirow{2}{*}{\textbf{ROUGE-1 (P)}} & \textbf{Gemini-1.5-Pro} & 0.347 & 0.322 & 0.321 \\ \cline{3-6} 
 &  & \textbf{GPT-4o} & 0.384 & 0.351 & 0.356 \\ \cline{2-6}
 & \multirow{2}{*}{\textbf{ROUGE-1 (R)}} & \textbf{Gemini-1.5-Pro} & 0.108 & 0.140 & 0.117 \\ \cline{3-6} 
 &  & \textbf{GPT-4o} & 0.073 & 0.142 & 0.148 \\ \cline{2-6}
 & \multirow{2}{*}{\textbf{ROUGE-1 (F1)}} & \textbf{Gemini-1.5-Pro} & 0.148 & 0.175 & 0.154 \\ \cline{3-6} 
 &  & \textbf{GPT-4o} & 0.151 & 0.247 & 0.251 \\ \cline{2-6}
 & \multirow{2}{*}{\textbf{ROUGE-L (P)}} & \textbf{Gemini-1.5-Pro} & 0.239 & 0.209 & 0.216 \\ \cline{3-6} 
 &  & \textbf{GPT-4o} & 0.274 & 0.232 & 0.227 \\ \cline{2-6}
 & \multirow{2}{*}{\textbf{ROUGE-L (R)}} & \textbf{Gemini-1.5-Pro} & 0.073 & 0.091 & 0.078 \\ \cline{3-6} 
 &  & \textbf{GPT-4o} & 0.073 & 0.142 & 0.148 \\ \cline{2-6}
 & \multirow{2}{*}{\textbf{ROUGE-L (F1)}} & \textbf{Gemini-1.5-Pro} & 0.100 & 0.113 & 0.102 \\ \cline{3-6} 
 &  & \textbf{GPT-4o} & 0.106 & 0.158 & 0.161 \\ \cline{2-6}
 & \multirow{2}{*}{\textbf{BERT Score (P)}} & \textbf{Gemini-1.5-Pro} & -0.127 & -0.111 & 0.120 \\ \cline{3-6} 
 &  & \textbf{GPT-4o} & -0.091 & -0.008 & 0.004 \\ \cline{2-6}
 & \multirow{2}{*}{\textbf{BERT Score (R)}} & \textbf{Gemini-1.5-Pro} & 0.028 & 0.032 & 0.010 \\ \cline{3-6} 
 &  & \textbf{GPT-4o} & 0.214 & 0.169 & 0.164 \\ \cline{2-6}
 & \multirow{2}{*}{\textbf{BERT Score (F1)}} & \textbf{Gemini-1.5-Pro} & -0.049 & -0.039 & -0.055 \\ \cline{3-6} 
 &  & \textbf{GPT-4o} & 0.057 & 0.080 & 0.084 \\ \hline
\multirow{18}{*}{\textbf{After Cut-Off}} & \multirow{2}{*}{\textbf{ROUGE-1 (P)}} & \textbf{Gemini-1.5-Pro} & 0.298 & 0.289 & 0.280 \\ \cline{3-6} 
 &  & \textbf{GPT-4o} & 0.383 & 0.335 & 0.328 \\ \cline{2-6}
 & \multirow{2}{*}{\textbf{ROUGE-1 (R)}} & \textbf{Gemini-1.5-Pro} & 0.126 & 0.154 & 0.140 \\ \cline{3-6} 
 &  & \textbf{GPT-4o} & 0.117 & 0.240 & 0.259 \\ \cline{2-6}
 & \multirow{2}{*}{\textbf{ROUGE-1 (F1)}} & \textbf{Gemini-1.5-Pro} & 0.161 & 0.185 & 0.171 \\ \cline{3-6} 
 &  & \textbf{GPT-4o} & 0.163 & 0.252 & 0.262 \\ \cline{2-6}
 & \multirow{2}{*}{\textbf{ROUGE-L (P)}} & \textbf{Gemini-1.5-Pro} & 0.201 & 0.187 & 0.185 \\ \cline{3-6} 
 &  & \textbf{GPT-4o} & 0.270 & 0.212 & 0.202 \\ \cline{2-6}
 & \multirow{2}{*}{\textbf{ROUGE-L (R)}} & \textbf{Gemini-1.5-Pro} & 0.080 & 0.101 & 0.092 \\ \cline{3-6} 
 &  & \textbf{GPT-4o} & 0.082 & 0.160 & 0.166 \\ \cline{2-6}
 & \multirow{2}{*}{\textbf{ROUGE-L (F1)}} & \textbf{Gemini-1.5-Pro} & 0.104 & 0.119 & 0.112 \\ \cline{3-6} 
 &  & \textbf{GPT-4o} & 0.113 & 0.162 & 0.163 \\ \cline{2-6}
 & \multirow{2}{*}{\textbf{BERT Score (P)}} & \textbf{Gemini-1.5-Pro} & -0.087 & -0.069 & -0.067 \\ \cline{3-6} 
 &  & \textbf{GPT-4o} & -0.034 & 0.050 & 0.062 \\ \cline{2-6}
 & \multirow{2}{*}{\textbf{BERT Score (R)}} & \textbf{Gemini-1.5-Pro} & -0.465 & -0.031 & -0.051 \\ \cline{3-6} 
 &  & \textbf{GPT-4o} & 0.241 & 0.181 & 0.174 \\ \cline{2-6}
 & \multirow{2}{*}{\textbf{BERT Score (F1)}} & \textbf{Gemini-1.5-Pro} & -0.066 & -0.049 & -0.058 \\ \cline{3-6} 
 &  & \textbf{GPT-4o} & 0.098 & 0.115 & 0.118 \\ \hline
\end{tabular}%
}
\end{table}

\vspace{-0.5em}

\onecolumn
\section{Prompts}
\subsection{Graphic Novels}
\label{appendix: E.1}

\begin{table}[h!]
\caption{The description of each prompt style is provided. We assigned a response format to FMs twice because, in Zero-Shot, the variation in responses is too broad, causing FMs to occasionally break the response format rule. In CoT+Zero-Shot, we utilized the simplest CoT style because it achieved the best score compared to the more detailed CoT version (Table 9.). In CoT+Few-Shot, we used three different examples. The performance was insufficient when using only one or two examples.}
\label{tab:prompts1}
\centering
\scriptsize 
\begin{tabular}{p{0.2\textwidth} p{0.75\textwidth}}
\toprule
\textbf{Graphic Novels} & \\ \midrule
\textbf{Example} & \textbf{Prompt} \\ \midrule
\textbf{Zero-Shot} &
\begin{tabular}[c]{@{}p{\textwidth}@{}}
Input : “The uploaded images represent parts of a story that has been shuffled and consists of 4 images."\\
"Arrange images in the correct order.”\\
“Respond with the list of numbers 1 to 4 in the following format only [1,2,3,4]”\\
“ONCE AGAIN!!! PLEASE!! respond with the list of numbers 1 to 4 in the following format only: [1,2,3,4]"\\
\\
(Task Images)\\
Output: A. \end{tabular}      \\\midrule

\textbf{CoT + Zero-Shot} &
\begin{tabular}[c]{@{}p{\textwidth}@{}}
Input : Q. “The uploaded images represent parts of a story that has been shuffled and consists of 4 images."\\ "Arrange images in the correct order.”\\
IMPORTANT: Respond ONLY with the list of numbers 1 to 4 in this format: [1, 2, 3, 4].\\
\\
(Task Images)\\
Output: A. Let's think step by step. The correct order is\\  \end{tabular} \\\midrule

\textbf{CoT + Few-Shot} &
\begin{tabular}[c]{@{}p{\textwidth}@{}}
Input : Q. “The uploaded images represent parts of a story that has been shuffled and consists of 4 images."\\ "Arrange images in the correct order.”\\
IMPORTANT: Respond ONLY with the list of numbers 1 to 4 in this format: [1, 2, 3, 4].\\
\\
\textbf{“The First, Example:”}:\\
\\
(1st Example Images)\\
A. “Let’s think step by step. The correct order is [1,2,3,4]”\\
\\
\textbf{“The Second, Example:”}:\\
\\
(2nd Example Images)\\ 
A. “Let’s think step by step. The correct order is [1,2,3,4]”\\
\\
\textbf{“The Third, Example:”}:\\
\\
(3rd Example Images)\\ 
A. “Let’s think step by step. The correct order is [1,2,3,4]”\\
\\
Q. “The uploaded images represent parts of a story that has been shuffled and consists of 4 images."\\ "Arrange images in the correct order.”\\
IMPORTANT: Respond ONLY with the list of numbers 1 to 4 in this format: [1, 2, 3, 4].\\
\\
(Task Images)\\
Output: A. Let’s think step by step. The correct order is\\ \end{tabular} \\\midrule
\end{tabular}
\end{table}

\clearpage
\captionsetup{justification=centering}
\subsection{Calligraphy}
\label{appendix: E.2}

\begin{table}[h!]
\caption{Korean Calligraphy Prompt: For the Cot+Few shot prompt, We utilized two examples but only one example is listed in the paper because it was too long to attach. The full prompt can be seen in GitHub.}
\label{tab:prompts2}
\centering
\scriptsize 
\begin{tabular}{p{0.2\textwidth} p{0.75\textwidth}}
\toprule
\textbf{Dataset Name} & \\ \midrule
\textbf{Example} & \textbf{Prompt} \\ \midrule
\textbf{Zero-Shot} &
\begin{tabular}[c]{@{}p{\textwidth}@{}}
Input : One korean calligraphy image \\\\
Prompt : "What are the all Korean characters in the image? \\Make sure that your answer only includes the result of the OCR without translating. \\You don't need to describe the processing steps." \\\\
Output: Only OCR result text
\end{tabular} \\\midrule

\textbf{CoT + Zero-Shot} &
\begin{tabular}[c]{@{}p{\textwidth}@{}}
Input : One korean calligraphy image \\\\
Prompt : "The image uploaded is Korean calligraphy with illustration. \\Transcribe the letters in the uploaded image. \\Solve it with following steps. \\1. Identify the start and end of the sentence. \\Check if there are any line breaks in the middle of the sentence. \\2. Split the recognized text into individual words. \\Combine the split words based on the context to form a coherent sentence. 
\\3. Analyze the context to infer the meaning of the handwriting. Correct typos by \\comparing them with similar words and choosing the correct one. \\4. Perform grammar and spelling checks to verify the recognized sentence.\\ Ensure that the sentence flows naturally and makes sense. \\Don't describe your steps. Just answer the result of the OCR without translating." \\\\
Output: Only OCR Result text
\end{tabular} \\\midrule

\textbf{CoT + Few-Shot} &
\begin{tabular}[c]{@{}p{\textwidth}@{}}
Input : One korean calligraphy image \\\\
Prompt : "Below are examples of OCR task. \\I'll show image first and explain step-by-step how to extract text from the image." \\\\
\textbf{Example1}: example1 image \\\\
"Step1: Identify the start and end of the sentence. Check if there are any line breaks in the middle of the sentence. \\Identify that the sentence starts with '바라는게' and ends with '안그래?' \\
Step2: Split into words and translate each word in English\\ and identify any typos based on the context.: 바라는게 (What I hope for) 무한정 (infinitely) 끝없이 (endlessly) 내리는 (falling) 게\\ (particle, indicating 'is') 아닌게 (is not) 엄마나 (Typo: misidentified word, Correct: 얼마나, Translation: how much)\\ 다행인지 (fortunately) 몰라 (I don't know) 안그래? (isn't it?) \\
Step3: Correct the typos by comparing each word with similar words\\ and combine the corrected words to form a coherent sentence.: \\'엄마나' should be '얼마나', '알고 래?' should be '안그래?' \\
Step4: Combine based on context:\\ '바라는게 무한정 끝없이 내리는 게 아닌게 얼마나 다행인지 몰라 안그래?' There is no weird word to use. \\
Step5: Analyze the context to infer the meaning of the handwriting. \\Correct any misrecognized words by comparing them with similar words and choosing the correct one. \\Infer the context: The sentence talks about how fortunate it is that something is not happening endlessly. \\Correct any misrecognized words: '얼 마나' should be '얼마나' \\
Step6: Perform grammar and spelling checks to verify the recognized sentence. \\Ensure that the sentence flows naturally and makes sense. Check grammar and spelling: \\Ensure '바라는게 무한정 끝없이 내리는 게 아닌게 얼마나 다행인지 몰라 안그래?' is grammatically correct and makes sense. \\Ensure the sentence flows naturally and the meaning is clear." \\
\\\\
prompt: "Now, please perform an OCR task on the following image like the example.\\ The image is Korean calligraphy with an illustration. \\Transcribe the letters in the picture with a step-by-step explanation of your reasoning. \\But Don't describe your steps. Just answer the result of the OCR without translating."\\\\
\\Output: Only OCR Result text
\end{tabular} \\\midrule
\end{tabular}
\end{table}

\clearpage

\subsection{Onion Not The Onion}
\label{appendix: E.3}
\begin{table}[h!]
\caption{We provided examples of prompts used to detect fake news, focusing on the implementation of CoT reasoning. We presented a structured approach that outlines the steps a FMs considers when analyzing and concluding whether a news story is fake or real. Lastly, this method involves a few-shot learning technique where examples of fake news and real news are given alongside rationales.}
\label{tab:fakenews_prompts}
\resizebox{\textwidth}{!}{%
\begin{tabular}{cl}
\toprule
{Onion, Not The Onion}\\
\\  \midrule
{Example} & {Prompt}\\
\\  \midrule \cline{1-2}
Zero Shot
& \begin{tabular}[c]{@{}l@{}}
\\
Input : A News article and Title \\
Prompt: The uploaded text is one of the articles that may be real or fake.\\ Please Answer whether below article is fake or real.\\ Say nothing but \textbf{the number 0 or 1}. i.e. Answer 1 if you think the article is real, \\answer 0 if you think it is fake\\
Output: (0 $\|$ 1) \end{tabular}      \\\hline

CoT + Zero Shot
& \begin{tabular}[c]{@{}l@{}}
\\
Input : A News article and Title \\
The uploaded text is one of the articles that may be real or fake.Please Answer whether below article is fake or real. \\Give a 20-character rationale for why you think that way, and output a 0 and 1 at the end of the sentence.\\To Solve this, You have to think step by step. \\\textbf{The first step} in identifying fake news is evaluating the reliability of the information source.\\ Well-known and verified news organizations are generally more reliable,\\ and their reports can be trusted more than unverified sources.\\ \textbf{In addition} to source reliability, look at the language used in the content.\\ Fake news often uses sensational or exaggerated language designed to elicit an emotional response. \\\textbf{It is also important to check} for consistency and accuracy in the information presented; \\fake news typically includes claims that are either unverified or clearly false.\\\textbf{Another critical step} is cross-verification,where check if the same claims are reported by multiple trusted sources.\\ i.e. rationale + answer 1 if you think the article is real, rationale + answer 0 if you think it is fake.\\ Must Keep in mind that the end of a sentence should end with either 0 or 1 \\ Output: (rationales + (0$\|$1)) \\ \end{tabular} \\\hline

CoT + few Shot
& \begin{tabular}[c]{@{}l@{}}
\\
Input : A News article and Title \\
The uploaded text is one of the articles that may be real or fake.Please Answer whether below article is fake or real. \\Give a 20-character rationale for why you think that way, and output a 0 and 1 at the end of the sentence.\\To Solve this, You have to think step by step. \\\textbf{The first step} in identifying fake news is evaluating the reliability of the information source.\\ Well-known and verified news organizations are generally more reliable,\\ and their reports can be trusted more than unverified sources.\\ \textbf{In addition} to source reliability, look at the language used in the content.\\ Fake news often uses sensational or exaggerated language designed to elicit an emotional response. \\\textbf{It is also important to check} for consistency and accuracy in the information presented; \\fake news typically includes claims that are either unverified or clearly false.\\\textbf{Another critical step} is cross-verification,where check if the same claims are reported by multiple trusted sources.\\\textbf{See the example below}. i.e. rationale + answer 1 if you think the article is real, rationale + answer 0 if you think it is fake. \\Must Keep in mind that the end of a sentence should end with either 0 or 1\\
\textbf{Example}: we provided one fake news story from The Onion and one real news story from Reddit's Not the Onion.\\ Additionally, rather than merely presenting the news, \\\textbf{we included examples of the rationales we derived for the two news stories, following the same prompting method}.\\
Output: (rationales + (0$\|$1)) \\ \end{tabular} \\
\\ \bottomrule \cline{1-2}
\end{tabular}
}
\end{table}

\clearpage 

\subsection{Lyrics}
\label{appendix: E.4}

\textbf{English Genre Classification}
\label{appendix: E.4.1}

\begin{table}[h!]
\caption{
Prompt for English genre classification task
}
\label{tab:lyrics_prompts_english_genre_task}
\centering
\small 
\begin{tabular}{p{3cm} p{13cm}} 
\toprule
{Lyrics} & \\ \midrule \hline
{Example} & {Prompt} \\ \midrule \cline{1-2}
Zero-Shot & \begin{tabular}[c]{@{}l@{}}
Input : Lyrics\\
\\
Prompt : Here is a list of unique music genres: ['genre list str'].\\
Say nothing but the Genre as Genre: the output. \\ 
Output example: Genre: [pop, r\&b, hip hop]. \\ 
Lyrics: 'lyrics'\\ 
\\
Output: Genre: {{the output}}\end{tabular} \\\hline

CoT + Zero-Shot & \begin{tabular}[c]{@{}l@{}}
Input : Lyrics \\
\\
Prompt : Here is a list of unique music genres: ['genre list str'].\\
Based on the lyrics provided, identify the genres.\\
Say nothing but the Genre as Genre: the output. \\ 
Output example: Genre: [pop, r\&b, hip hop]. \\ 
Lyrics: 'lyrics'\\ 
\\ 
Output: Genre: {{the output}} \\ \end{tabular} \\\hline

CoT + Few-Shot & \begin{tabular}[c]{@{}l@{}}
Input : Lyrics\\
Prompt : 
Here is a list of unique music genres: ['genre list str'].\\

\\ \textbf{Example Lyrics}:\\     
And she spoke words that would melt in your hands\\
And she spoke words of wisdom\\
To the basement, people, to the basement\\
Many surprises await you\\
In the basement, people, in the basement\\

You hid there last time, you know we're gonna find you\\
Sick in the car seat, 'cause you're not up to going\\
Out on the main streets, completing your mission\\
You hid there last time, you know we're gonna find you\\
Sick in the car seat, 'cause you're not up to going\\
Out on the main streets, completing your mission\\

\\ \textbf{Example Description}: indie pop\\
\\
Now, based on the lyrics provided, identify the genres.\\
Say nothing but the Genre as Genre: the output. \\ 
Output example: Genre: [pop, r\&b, hip hop]. \\ 
Lyrics: 'lyrics'\\ 
\\ 
Output: Genre: {{the output}} \\ \end{tabular} \\
\\ \bottomrule
\end{tabular}
\end{table}

\clearpage

\textbf{Korean Genre Classification}
\label{appendix: E.4.2}

\begin{table}[h!]
\caption{
Prompt for Korean genre classification task
}
\label{tab:lyrics_prompts_korean_genre_task}
\centering
\small 
\begin{tabular}{p{3cm} p{13cm}} 
\toprule
{Lyrics} & \\ \midrule \hline
{Example} & {Prompt} \\ \midrule \hline
Zero-Shot
& \begin{tabular}[c]{@{}l@{}}
Input : Lyrics\\
\\
Prompt : Here is a list of unique music genres: ['genre list str'].\\
Say nothing but the Genre as Genre: the output. \\ 
Output example: Genre: [발라드, 댄스, 랩/힙합]. \\ 
Lyrics: 'lyrics'\\ 
 \\
Output: Genre: {{the output}}\end{tabular} \\\hline

CoT + Zero-Shot
& \begin{tabular}[c]{@{}l@{}}
Input : Lyrics \\
\\
Prompt : Here is a list of unique music genres: ['genre list str'].\\
Based on the lyrics provided, identify the genres.\\
Say nothing but the Genre as Genre: the output. \\ 
Output example: Genre: [발라드, 댄스, 랩/힙합]. \\ 
Lyrics: 'lyrics'\\ 
\\ 
Output: Genre: {{the output}} \\ \end{tabular} \\\hline

CoT + Few-Shot
& \begin{tabular}[c]{@{}l@{}}
Input : Lyrics\\
Prompt : 
Here is a list of unique music genres: ['genre list str'].\\

\\ \textbf{Example Lyrics}:\\     
처음 그대 내게로 오던 그날에\\
잠시 동안 적시는\\
그런 비가 아니길\\
간절히 난 바래왔었죠\\
그대도 내 맘 아나요\\
매일 그대만 그려왔던 나를\\
오늘도 내 맘에 스며들죠\\
그대는 선물입니다\\
하늘이 내려준\\
홀로 선 세상 속에\\
그댈 지켜줄게요\\
어느 날 문득\\
소나기처럼\\
내린 그대지만\\
오늘도 불러 봅니다\\
내겐 소중한 사람\\
Oh\\
떨어지는 빗물이\\
어느새 날 깨우고\\
그대 생각에 잠겨요\\
이제는 내게로 와요\\
언제나처럼 기다리고 있죠\\
그대 손을 꼭 잡아줄게요'\\

\\ \textbf{Example Description}: 발라드, 국내드라마\\
\\
Now, based on the lyrics provided, identify the genres.\\
Say nothing but the Genre as Genre: the output. \\ 
Output example: Genre: [발라드, 댄스, 랩/힙합]. \\ 
Lyrics: 'lyrics'\\ 
\\ 
Output: Genre: {{the output}} \\ \end{tabular} \\
\\ \bottomrule
\end{tabular}

\end{table}

\clearpage

\textbf{English Song Description Generation}
\label{appendix: E.4.3}

\begin{table}[h!]
\caption{
Prompt for English song description generation task
}
\label{tab:lyrics_prompts_english_description_task}
\resizebox{\textwidth}{!}{%
\centering
\begin{tabular}{cl}
\toprule
{Lyrics}\\
\\  \midrule \hline
{Example} & {Prompt}\\
\\  \midrule \cline{1-2}
Zero-Shot
& \begin{tabular}[c]{@{}l@{}}
\\
Input : Lyrics\\
\\
Prompt : Say nothing but the Description as Description: {the output}\\

Output example: Description: The song explores themes of love and heartbreak.\\

Lyrics: '{lyrics}'\\
 \\
Output: Description: {{the output}}\end{tabular}      \\\hline

CoT + Zero-Shot
& \begin{tabular}[c]{@{}l@{}}
\\
Input : Lyrics \\
\\
Prompt : Based on the provided lyrics, write a brief description of the song.\\

Include the possible song title and artist name in the description.\\

Say nothing but the Description as Description: {the output}\\

Output example: Description: Honeymoon Avenue by Ariana Grande is about knowing you are at the end of a relationship  \\ and wishing it could not be the end and go back to the beginning and start over. \\
\\ 
Output: Description: {{the output}} \\ \end{tabular} \\\hline

CoT + Few-Shot
& \begin{tabular}[c]{@{}l@{}}
\\
Input : Lyrics\\
Prompt : 
\\ \textbf{Example Lyrics}: I'd like to say we gave it a try\\
I'd like to blame it all on life\\
Maybe we just weren't right\\
But that's a lie, that's a lie\\

And we can deny it as much as we want\\
But in time, our feelings will show\\
'Cause sooner or later, we'll wonder why we gave up\\
The truth is everyone knows, oh\\

Almost, almost is never enough\\
So close to being in love\\
If I would have known that you wanted me the way I wanted you\\
Then maybe we wouldn't be two worlds apart (Ah)\\
But right here in each other's arms\\
And we almost, we almost knew what love was\\
But almost is never enough (Ah)\\

If I could change the world overnight (Ah)\\
There'd be no such thing as goodbye (Ah)\\
You'd be standing right where you were (Ah)\\
And we'd get the chance we deserve, oh (Ah)\\
See upcoming pop shows\\
Get tickets for your favorite artists\\

Try to deny it as much as you want\\
But in time, our feelings will show (Ah)\\
'Cause sooner or later, we'll wonder why we gave up\\
The truth is everyone knows (Ah)\\

\\
\\ \textbf{Example Description}: On the collaborative track “Almost Is Never Enough,” Ariana Grande \& Nathan Sykes play a couple who had a relationship that hadn’t gone right.\\
Ariana would like to say things were going well but she knows that’s a lie and like the title states, almost is never enough to make the relationship work; you need to put full effort in.\\
Both of them state that they didn’t feel the relationship while in it, but the mood of the song and lyrics suggest that they both want to either reconnect or they simply just miss better times.\\

At the time of the song’s release, Nathan and Ariana were dating. Unfortunately, their relationship ended a few months later.
\\ \\
Now, based on the provided lyrics, write a brief description of the song.\\

Include the possible song title and artist name in the description.\\

Say nothing but the Description as Description: {the output}\\

Output example: Description: Honeymoon Avenue by Ariana Grande is about knowing you are at the end of a relationship \\ and wishing it could not be the end and go back to the beginning and start over. \\

\\ 

Output: Description: {{the output}}\\ \end{tabular} \\
\\ \bottomrule \cline{1-2}
\end{tabular}
}

\end{table}

\clearpage

\textbf{English Song Infilling}
\label{appendix: E.4.4}

\begin{table}[h!]
\caption{
Prompt for English lyrics infilling task. Examples in CoT+Few-shot are composed of data removed during BERT testing.
}
\label{tab:lyrics_prompts_english_infilling_task}
\scriptsize 
\resizebox{\textwidth}{!}{%
\centering
\begin{tabular}{cl}
\toprule
{Lyrics Infilling Task}\\
\\  \midrule \hline
{Example} & {Prompt}\\
\\  \midrule \cline{1-2}
Zero-Shot
& \begin{tabular}[c]{@{}l@{}}
\\
Input : Masked lyrics\\
\\
Prompt : You are a powerful language model. Fill in the blanks in the following text with appropriate words.\\The text is a part of a song with certain words masked by [MASK]. \\Lyrics: '{lyrics} \\Say nothing but the filled lyrics as 'Filled lyrics: {{the output}}'. \\Output example: Filled lyrics: 'I know this pain (I know this pain) why do you lock yourself up in these chains? (these chains)…\\
\\
Output: Filled lyrics: {{the output}}\end{tabular}      \\\hline

CoT + Zero-Shot
& \begin{tabular}[c]{@{}l@{}}
\\
Input : Lyrics \\
\\
Prompt : You are a powerful language model. Fill in the blanks in the following text with appropriate words.The text is a part of a song with certain words masked by [MASK]. \\For each blank, think step by step about the context and meaning of the surrounding text before choosing the word. \\To do this, follow these steps: \\  a. Carefully read and analysis the lyrics. \\  b-1. Check the entire lyrics to see if there are any repeating parts. \\  b-2. If repeating parts exist, replace the [MASK] with the corresponding word. \\   c-1. Make the list of possible words for the masked part. \\  c-2. Select a suitable word from the candidate list. \\  c-3. Replace [MASK] with the word that you selected.\\Lyrics: '{lyrics}\\Step-by-step reasoning and filled lyrics as 'Filled lyrics: {{the output}}'.\\Say nothing but the filled lyrics as 'Filled lyrics: {{the output}}'. \\Output example: Filled lyrics: 'I know this pain (I know this pain) why do you lock yourself up in these chains? (these chains)…\\

 \\
Output: Filled lyrics: {{the output}} \\ \end{tabular} \\\hline

CoT + Few-Shot
& \begin{tabular}[c]{@{}l@{}}
\\
Input : Lyrics\\
Prompt :\\

You are a powerful language model. Fill in the blanks in the following text with appropriate words. The text is a part of a song with certain words masked by [MASK]. \\For each blank, think step by step about the context and meaning of the surrounding text before choosing the word. \\To do this, follow these steps: \\  a. Carefully read and analysis the lyrics. \\  b-1. Check the entire lyrics to see if there are any repeating parts. \\  b-2. If repeating parts exist, replace the [MASK] with the corresponding word. \\   c-1. Make the list of possible words for the masked part. \\  c-2. Select a suitable word from the candidate list. \\  c-3. Replace [MASK] with the word that you selected. \\ \\Example: \\Lyrics: \\Rotgut whiskey's gonna ease my mind Beach [MASK] rests on the dryin' line \\Do I remind you of your daddy in his '88 Ford? Labrador [MASK] out the passenger door \\The sand from your hair is blowin' in my eyes [MASK] it on [MASK] [MASK] grown men \\don't cry [MASK] [MASK] remember that beat down basement couch? \\I'd sing [MASK] my love songs [MASK] you'd tell me about \\How your mama [MASK] off and pawned her ring [MASK] remember, \\I remember everything \\Filled lyrics: \\Rotgut whiskey's gonna ease my mind Beach towel rests on the dryin' line \\Do I remind you of your daddy in his '88 Ford? Labrador hangin' out the passenger door \\The sand from your hair is blowin' in my eyes Blame it on the beach, grown men \\don't cry Do you remember that beat down basement couch? \\I'd sing you my love songs and you'd tell me about \\How your mama ran off and pawned her ring I remember, \\I remember everything \\ \\Now, based on the provided lyrics, fill in the blanks with appropriate words. \\Lyrics: '{lyrics}\\Step-by-step reasoning and filled lyrics as 'Filled lyrics: {{the output}}'.\\Say nothing but the filled lyrics as 'Filled lyrics: {{the output}}'. \\Output example: Filled lyrics: 'I know this pain (I know this pain) why do you lock yourself up in these chains? (these chains)…\\ 

Output: Filled lyrics: {{the output}}\\ \end{tabular} \\
\\ \bottomrule \cline{1-2}
\end{tabular}
}
\end{table}

\clearpage

\textbf{Korean Song Infilling task}
\label{appendix: E.4.5}

\begin{table}[h!]
\caption{
Prompt for Korean lyrics infilling task. Examples in CoT+Few-shot are composed of data removed during BERT testing.
}
\label{tab:lyrics_prompts_korean_infilling_task}
\resizebox{\textwidth}{!}{%
\centering
\begin{tabular}{cl}
\toprule
{Lyrics}\\
\\  \midrule \hline
{Example} & {Prompt}\\
\\  \midrule \cline{1-2}
Zero-Shot
& \begin{tabular}[c]{@{}l@{}}
\\
Input : Masked lyrics\\
\\
Prompt : You are a powerful language model. Fill in the blanks in the following text with appropriate words.\\The text is a part of a song with certain words masked by [MASK]. \\Lyrics: '{lyrics} \\Say nothing but the filled lyrics as 'Filled lyrics: {{the output}}'. \\Output example: Filled lyrics: 'I know this pain (I know this pain) why do you lock yourself up in these chains? (these chains)…\\
\\
Output: Filled lyrics: {{the output}}\end{tabular}      \\\hline

CoT + Zero-Shot
& \begin{tabular}[c]{@{}l@{}}
\\
Input : Lyrics \\
\\
Prompt : You are a powerful language model. Fill in the blanks in the following text with appropriate words.\\The text is a part of a song with certain words masked by [MASK]. \\For each blank, think step by step about the context and meaning of the surrounding text before choosing the word. \\To do this, follow these steps: \\  a. Carefully read and analysis the lyrics. \\  b-1. Check the entire lyrics to see if there are any repeating parts. \\  b-2. If repeating parts exist, replace the [MASK] with the corresponding word. \\   c-1. Make the list of possible words for the masked part. \\  c-2. Select a suitable word from the candidate list. \\  c-3. Replace [MASK] with the word that you selected.\\Lyrics: '{lyrics}\\Step-by-step reasoning and filled lyrics as 'Filled lyrics: {{the output}}'.\\Say nothing but the filled lyrics as 'Filled lyrics: {{the output}}'. \\Output example: Filled lyrics: 'I know this pain (I know this pain) why do you lock yourself up in these chains? (these chains)…\\

 \\
Output: Filled lyrics: {{the output}} \\ \end{tabular} \\\hline

CoT + Few-Shot
& \begin{tabular}[c]{@{}l@{}}
\\
Input : Lyrics\\
Prompt :\\

You are a powerful language model. Fill in the blanks in the following text with appropriate words.\\The text is a part of a song with certain words masked by [MASK]. \\For each blank, think step by step about the context and meaning of the surrounding text before choosing the word. \\To do this, follow these steps: \\  a. Carefully read and analysis the lyrics. \\  b-1. Check the entire lyrics to see if there are any repeating parts. \\  b-2. If repeating parts exist, replace the [MASK] with the corresponding word. \\   c-1. Make the list of possible words for the masked part. \\  c-2. Select a suitable word from the candidate list. \\  c-3. Replace [MASK] with the word that you selected. \\ \\Example: \\Lyrics: \\세상에 음악의 신이 있다면 고맙다고 안아주고 싶어 전 세계 공통의 Language 자음과 모음이 달라도 상관없는 건 Music\\말이 안 통해도 [MASK] 있다면 [MASK] 지금부터는 아주 친한 친구 너와 내가 모르는 사이여도 춤출 [MASK] 있어 We [MASK] mix it up right\\Sugar and spice Brass sound and guitar 네 [MASK] 다 내 [MASK] 쿵치팍치 또한 내 이름인가\\이것 또한 나를 위한 소린가 [MASK] [MASK] Drum bass Piano [MASK]\\Filled lyrics: \\세상에 음악의 신이 있다면 고맙다고 안아주고 싶어  전 세계 공통의 Language 자음과 모음이 달라도 상관없는 건 Music\\말이 안 통해도 음악이 있다면 우리는 지금부터는 아주 친한 친구  너와 내가 모르는 사이여도 춤출 수 있어 We can mix it up right\\Sugar and spice Brass sound and guitar  네 글자면 다 내 이름이래 쿵치팍치 또한 내 이름인가\\이것 또한 나를 위한 소린가 Kick snare Drum bass Piano Bassline\\ \\Lyrics: '{lyrics}\\Step-by-step reasoning and filled lyrics as 'Filled lyrics: {{the output}}'.\\Say nothing but the filled lyrics as 'Filled lyrics: {{the output}}'. \\Output example: Filled lyrics: 'I know this pain (I know this pain) why do you lock yourself up in these chains? (these chains)…\\ 

Output: Filled lyrics: {{the output}}\\ \end{tabular} \\
\\ \bottomrule \cline{1-2}
\end{tabular}
}

\end{table}

\end{document}